\documentclass[10pt,twocolumn,letterpaper]{article}

\usepackage[pagenumbers]{arxiv_main} % To force page numbers, e.g. for an arXiv version

\usepackage{setspace}

\usepackage[dvipsnames]{xcolor}
\usepackage{mathtools}
\usepackage{listings}
\usepackage{float}
\usepackage{arxiv_main}
\usepackage{times}
\usepackage{epsfig}
\usepackage{graphicx}
\usepackage{amsmath}
\usepackage{amssymb}
\usepackage{booktabs}
\usepackage{bm}
\usepackage{float}
\usepackage{algorithm}
\usepackage{algorithmic}

\usepackage[accsupp]{axessibility}  % Improves PDF readability for those with disabilities.

\usepackage{multicol}
\usepackage{multirow}
\usepackage{caption}
\usepackage{graphicx}
\usepackage{adjustbox}

\usepackage{array}
\newcolumntype{P}[1]{>{\centering\arraybackslash}p{#1}}
\usepackage{dblfloatfix}
\usepackage{enumitem}
\usepackage{mathabx}

\newcommand\mycommfont[1]{\footnotesize\ttfamily\textcolor{gray}{#1}}
\renewcommand\ttdefault{cmvtt}
\newcommand{\ba}{\mathbf{a}}
\newcommand{\bb}{\mathbf{b}}
\newcommand{\bc}{\mathbf{c}}
\newcommand{\bh}{\mathbf{h}}
\newcommand{\bk}{\mathbf{k}}
\newcommand{\bo}{\mathbf{o}}
\newcommand{\bs}{\mathbf{s}}
\newcommand{\bt}{\mathbf{t}}
\newcommand{\bu}{\mathbf{u}}
\newcommand{\bv}{\mathbf{v}}
\newcommand{\bw}{\mathbf{w}}
\newcommand{\bx}{\mathbf{x}}
\newcommand{\by}{\mathbf{y}}
\newcommand{\bz}{\mathbf{z}}

\newcommand{\bA}{\mathbf{A}}
\newcommand{\bB}{\mathbf{B}}
\newcommand{\bC}{\mathbf{C}}
\newcommand{\bD}{\mathbf{D}}
\newcommand{\bE}{\mathbf{E}}
\newcommand{\bF}{\mathbf{F}}
\newcommand{\bG}{\mathbf{G}}
\newcommand{\bH}{\mathbf{H}}
\newcommand{\bI}{\mathbf{I}}
\newcommand{\bL}{\mathbf{L}}
\newcommand{\bO}{\mathbf{O}}
\newcommand{\bR}{\mathbf{R}}
\newcommand{\bS}{\mathbf{S}}
\newcommand{\bT}{\mathbf{T}}
\newcommand{\bU}{\mathbf{U}}
\newcommand{\bV}{\mathbf{V}}
\newcommand{\bW}{\mathbf{W}}
\newcommand{\bX}{\mathbf{X}}
\newcommand{\bY}{\mathbf{Y}}
\newcommand{\bZ}{\mathbf{Z}}

\newcommand{\bOnes}{\mathbf{1}}
\newcommand{\bZeros}{\mathbf{0}}
\newcommand{\bTheta}{\mathbf{\Theta}}

\newcommand{\bsh}{\boldsymbol{h}}
\newcommand{\bsi}{{\boldsymbol{i}}}
\newcommand{\bst}{{\boldsymbol{t}}}
\newcommand{\bsu}{{\boldsymbol{u}}}
\newcommand{\bsz}{{\boldsymbol{z}}}

\newcommand{\bsmu}{\boldsymbol{\mu}}
\newcommand{\bsmui}{\boldsymbol{\mu}^i}
\newcommand{\bsmut}{\boldsymbol{\mu}^t}

\newcommand{\bLm}{\mathbf{L}^m}

\newcommand{\bhi}{\mathbf{h}^{{i}}}
\newcommand{\bht}{\mathbf{h}^{{t}}}
\newcommand{\bxi}{\mathbf{x}^{{i}}}
\newcommand{\bxt}{\mathbf{x}^{{t}}}

\newcommand{\bXi}{\mathbf{X}^{{i}}}
\newcommand{\bXt}{\mathbf{X}^{{t}}}

\newcommand{\bshi}{\boldsymbol{h}^{{i}}}
\newcommand{\bsht}{\boldsymbol{h}^{{t}}}
\newcommand{\bsxi}{\boldsymbol{x}^{{i}}}
\newcommand{\bsxt}{\boldsymbol{x}^{{t}}}

\newcommand{\transpose}{\hspace{-0.15em}^\top\hspace{-0.15em}}
\newcommand{\eq}{\hspace{-0.15em}=\hspace{-0.15em}}

\newcommand{\bbE}{\mathbb{E}}
\newcommand{\bbH}{\mathbb{H}}
\newcommand{\bbI}{\mathbb{I}}
\newcommand{\bbJ}{\mathbb{J}}
\newcommand{\bbM}{\mathbb{M}}
\newcommand{\bbR}{\mathbb{R}}

\newcommand{\ndis}{\mathcal{N}}
\newcommand{\xset}{\mathcal{X}}
\newcommand{\oset}{\mathcal{O}}
\newcommand{\vset}{\mathcal{V}}
\newcommand{\tset}{\mathcal{T}}
\newcommand{\uset}{\mathcal{U}}
\newcommand{\dis}{\mathcal{D}}
\newcommand{\func}{\boldsymbol{f}}
\newcommand{\tr}{\text{Tr}}
\newcommand{\m}{{(m)}}
\newcommand{\mt}{{(t)}}
\newcommand{\1}{{(1)}}
\newcommand{\2}{{(2)}}
\newcommand{\3}{{(3)}}
\newcommand{\4}{{(4)}}
\newcommand{\5}{{(5)}}
\newcommand{\M}{{(M)}}
\newcommand{\sign}{\mathrm{sign}}
\newcommand{\diag}{\textrm{diag}}
\usepackage{amssymb}% http://ctan.org/pkg/amssymb
\usepackage{pifont}% http://ctan.org/pkg/pifont
\newcommand{\cmark}{\ding{51}}%
\newcommand{\xmark}{\ding{55}}%

\newcommand{\IJS}{I_{\text{JS}}}
\newcommand{\DJS}{D_{\text{JS}}\hspace{-1pt}}
\newcommand{\DSKL}{D_{\text{SKL}}\hspace{-1pt}}
\newcommand{\DKL}{D_{\text{KL}}\hspace{-1pt}}
\usepackage[pagebackref,breaklinks,colorlinks]{hyperref}

\usepackage[capitalize]{cleveref}
\crefname{section}{Sec.}{Secs.}
\Crefname{section}{Section}{Sections}
\Crefname{table}{Table}{Tables}
\crefname{table}{Tab.}{Tabs.}

\def\cvprPaperID{1171} % *** Enter the CVPR Paper ID here
\def\confName{CVPR}
\def\confYear{2023}

\newcommand{\chao}[1]{\textcolor{purple}{CD: #1}}
\newcommand{\tianyu}[1]{\textcolor{purple}{TY: #1}}

\newcommand{\nocontentsline}[3]{}
\newcommand{\tocless}[2]{\bgroup\let\addcontentsline=\nocontentsline#1{#2}\egroup}
\begin{document}


\title{
% On Transfer Learning for
% Alleviating 
% Discovering
Exploring Incompatible Knowledge Transfer in 
Few-shot Image Generation
}

\author{
\vspace{-2mm}
Yunqing Zhao$^{1}$
\and
Chao Du$^{2}$
\and
Milad Abdollahzadeh$^{1}$
\and
Tianyu Pang$^{2}$
\and
\vspace{-4mm}
\\
Min Lin$^{2}$\hspace{3mm}
Shuicheng Yan$^{2}$\hspace{3mm}
Ngai-Man Cheung$^{1}$
\vspace{1mm}
\\
{
$^{1}$ Singapore University of Technology and Design (SUTD) \hspace{3.5mm}
$^{2}$ Sea AI Lab \hspace{3.5mm}
}
\\
{
\tt\small yunqing\_zhao@mymail.sutd.edu.sg,
}
{
\tt\small ngaiman\_cheung@sutd.edu.sg
}
}
\maketitle

% ---------------------------- Main Text ---------------------------- %
% ------------------------------------------------------------------- %

% Abstract
%%%%%%%%% ABSTRACT
\begin{abstract}
\vspace{-0.1cm}
Few-shot image generation (FSIG) learns to generate diverse and high-fidelity images from a target domain using a few (\eg, 10) reference samples. Existing FSIG methods select, preserve and transfer prior knowledge from a source generator (pretrained on a related domain) to learn the target generator. In this work, we investigate an underexplored issue in FSIG, dubbed as incompatible knowledge transfer, which would significantly degrade the realisticness of synthetic samples. Empirical observations show that the issue stems from the least significant filters from the source generator. To this end, we propose knowledge truncation to mitigate this issue in FSIG, which is a complementary operation to knowledge preservation and is implemented by a lightweight pruning-based method. Extensive experiments show that knowledge truncation is simple and effective, consistently achieving state-of-the-art performance, including challenging setups where the source and target domains are more distant.
% with different GAN architectures. 
{Project Page:} {\color{RubineRed}\href{https://yunqing-me.github.io/RICK}{yunqing-me.github.io/RICK}}.
\vspace{-0.5cm}
\end{abstract}

% Introduction
\tocless

\section{Introduction}
\label{sec-1}

% % ---------------------------------------------- %
% % ---------------------------------------------- %

Over recent years, 
deep generative models~\cite{kingma2014adam,goodfellow2014GAN,rezende2015variational,ho2020denoising, gong2022diffpose, Zhao_2023_arxiv_watermark_dm} have made tremendous progress, enabling many intriguing tasks such as image generation~\cite{brock2018bigGAN, karras2018styleGANv1, karras2020styleganv2, Karras2021styleganv3}, image editing~\cite{lin2021anycostGANs, yanbo_2022_CVPR_editing, zhu2020domain, wang2022high}, and data augmentation~\cite{adar:liver:2018,tran2021data_aug_gan, chai2021ensembling}.
In spite of their remarkable success,
most research on generative models has been focusing on setups with sizeable training datasets~\cite{karras2020ADA, tran2021data_aug_gan, zhao2020leveraging_icml_adafm, tseng2021regularizingGAN, yang2021InsGen, huang2022masked},
limiting its applications in many domains where data collection is difficult or expensive~\cite{adar:liver:2018, gong2022system, gong2022meta} (\eg, medicine).
% However, in many critical domains (\eg, medical), collection of data points is challenging and expensive \cite{adar:liver:2018}.

To address such problems, FSIG has been proposed recently~\cite{li2020fig_EWC, ojha2021fig_cdc}, which learns to generate images with extremely few reference samples (\eg, 10 samples) from a target domain. In these regimes, learning a generator to capture the underlying target distribution is an undetermined problem that requires some prior knowledge. The majority of existing state-of-the-art (SOTA) FSIG methods rely on transfer learning approaches~\cite{wang2018transferringGAN,Yosinski:transfer:nips14, Hu_2022_CVPR_few_shot_transfer, Afrasiyabi_2022_CVPR_few_shot_transfer} to exploit prior knowledge (\eg, a source generator) learned from abundant data of a different but related source domain, and then transfer suitable source knowledge to learn the target generator with fine-tuning~\cite{wang2018transferringGAN, noguchi2019BSA, mo2020freezeD, karras2020ADA, cong2020gan_memory, wang2020minegan, li2020fig_EWC, zhao2020leveraging_icml_adafm, ojha2021fig_cdc, zhao2022dcl, zhao2022fsig-ip}. Different techniques have been proposed to effectively preserve useful source knowledge, such as freezing \cite{mo2020freezeD}, regularization \cite{li2020fig_EWC, ojha2021fig_cdc, zhao2022dcl} and modulation \cite{zhao2022fsig-ip} (details in Sec. \ref{sec-2}).

\textbf{Incompatible knowledge transfer.}
Despite the impressive improvement achieved by different {knowledge preservation} approaches~\cite{mo2020freezeD,li2020fig_EWC,ojha2021fig_cdc,zhao2022dcl,zhao2022fsig-ip}, in this work, we argue that preventing \emph{incompatible knowledge transfer} is equally crucial. This is revealed through a carefully designed investigation, where such {incompatible knowledge transfer} is manifested in the presence of unexpected semantic features. These features are inconsistent with the target domain, thereby degrading the realisticness of synthetic samples. As illustrated in Figure~\ref{bad_knowledge_transfer}, trees and buildings are incompatible with the domain of {\small\texttt{Sailboat}} (as can be observed by inspecting the 10 reference samples).
However, they appear in the synthetic images when applying the existing SOTA methods~\cite{li2020fig_EWC, zhao2022fsig-ip} with a source generator trained on {\small\texttt{Church}}. This shows that the existing methods cannot effectively prevent the transfer of incompatible knowledge.
\textbf{Knowledge truncation.}
Based on our observations, we propose {\bf R}emoving {\bf I}n-{\bf C}ompatible {\bf K}nowledge (RICK), a lightweight filter-pruning based method to remove filters that encode incompatible knowledge (\ie, filters with least estimated importance for adaptation) during FSIG adaptation.  
 While filter pruning has been applied extensively to achieve compact deep networks with reduced computation \cite{he2019filter, sui2021chip, prune_redundancy}, its application to prevent transfer of incompatible knowledge is underexplored. We note that our proposed knowledge truncation and pruning of incompatible filters are orthogonal and complementary 
 with existing knowledge preservation methods in FSIG. 
 %As a result,
 In this way, our method effectively removes the incompatible knowledge compared to prior works, and 
 achieves noticeably improved quality (\eg, FID \cite{heusel2017FID}) of generated images.

 Our contributions can be summarized as follows:
 
%  $\bullet$ We revisit recent SOTA methods for FSIG and uncover the important issue that incompatible features are still \textit{transferred} to the target generator after few-shot adaptation, and this is due to imperfect \textit{knowledge update}.

%  {\hspace{-4 mm}}
 $\bullet$ We explore the incompatible knowledge transfer for FSIG, reveal that SOTA methods fail in handling this issue, investigate the underlying causation, and %expose 
 disclose the inadequacy of fine-tuning in removal of incompatible knowledge.

%  {\hspace{-4 mm}}
%  $\bullet$ We propose knowledge truncation,  a complementary operation for FSIG to alleviate incompatible knowledge transfer. 
%  As a first realisation, we propose a lightweight filter-pruning based method for knowledge truncation.
 $\bullet$ We propose knowledge truncation to alleviate incompatible knowledge transfer, and realize it with a lightweight filter-pruning based method.
 
%  {\hspace{-4 mm}}
%  $\bullet$ Extensive experiments show that our method effectively removes incompatible knowledge and consistently improves the quality of generated images, including challenging setups where source and target domains are dissimilar.
 $\bullet$ Extensive experiments show that our method effectively removes incompatible knowledge and consistently improves the generative quality, including challenging setups where source and target domains are dissimilar.
\tocless
\section{Related Works}
\label{sec-2}

% In this paper, we perform an in-depth study of transfer learning and knowledge transfer for FSIG. Transfer learning 

%{\bf Transfer learning in FSIG.}
Transfer learning (TL) \cite{pan2009yang-qiang-transfer,zhuang:transfer:IEEE2021, lihonglin2023task} is a widely used approach to improve the performance of a model  
in a data-limited target domain 
by leveraging the knowledge of a model pretrained on a data-rich source domain \cite{deng2009imagenet, zhou2017places, li_new2021uav}.
Conventionally, TL has been applied to {predictors \cite{li_new2020hard, li_new2021else, li_new2022dynamic}}, including image classifiers \cite{Yosinski:transfer:nips14, Hu_2022_CVPR_few_shot_transfer, Afrasiyabi_2022_CVPR_few_shot_transfer, Zhao_2023_tip_fsc, sun2020explanation, keshik2022ls-kd}, or object detectors \cite{Sermanet:Overfeat:iclr2014, Gupta_2022_CVPR_detection_transfer, li2022source}.
The primary focus of TL has been on selecting and preserving useful knowledge of the source model into the target model \cite{pan2009yang-qiang-transfer,zhuang:transfer:IEEE2021}.
For example, \cite{Fergus:visualize:2013,Sermanet:Overfeat:iclr2014, li:inductive:icml18, percy:finetuning:iclr22}
preserve and transfer generalizable layers of a source network into the target network. 
% Li \etal. \cite{li:inductive:icml18} investigated several regularization schemes that retain source knowledge into the target model. 
Recent work searches for useful features in the entire source network, preserves them in the target network, and trains a linear classification head on top of the preserved features \cite{Evci:head2toe:icml2022}.

Besides predictive models, TL has been applied to {generative models} recently for FSIG
\cite{mo2020freezeD,li_new2021pot, ojha2021fig_cdc,zhao2022dcl,zhao2022fsig-ip},
% %  \textbf{Transfer learning for FSIG.} 
%  Different from discriminative few-shot learning, \eg, few-shot classification \cite{snell2017prototypical, xie2022joint}, segmentation \cite{lang2022learning_what_not_to_seg, wang2022remember} or detection \cite{han2022few, kaul2022label}, where the abundant training data is available, previous works of FSIG often adopts transfer learning approach \cite{pan2009yang-qiang-transfer}, 
 where they adapt a GAN pretrained on a large source dataset as initialization, and perform adaptation on very limited target training domains. 
 For example, baseline methods such as TGAN \cite{wang2018transferringGAN} simply fine-tune the pretrained model via GAN loss (see Eqn.~(\ref{ladv})).
 Recent state-of-the-art methods propose to preserve some knowledge for adaptation. For example,
 FreezeD \cite{mo2020freezeD} fixes some low-level layers of the discriminator for adaptation; 
 EWC \cite{li2020fig_EWC} identifies the important parameter of a source task and penalizes the weights change; CDC \cite{ojha2021fig_cdc} aims to preserve the consistency of distance between generated images before and after adaptation; DCL \cite{zhao2022dcl} maximizes the mutual information between generated images on source and target from the same input latent code to preserve the knowledge.
 More recently, AdAM \cite{zhao2022fsig-ip} proposes a modulation based method to identify the source knowledge important for the target domain, and preserves the knowledge for adaptation.

\tocless

% --------------------- Preliminary ----------------------- %
\section{Preliminaries}
\label{sec-prelim}
Existing FSIG methods adopt TL approach and leverage a source GAN pretrained on a large source dataset. We denote the source generator as $G_{s}$ (source discriminator as $D_{s}$). 
During adaptation, the target generator $G_{t}$ (target discriminator as $D_{t}$) is obtained by {\em fine-tuning} the source GAN on few-shot target images via adversarial loss
 $\mathcal{L}_{adv}$ \cite{goodfellow2014GAN}: 
\vspace{-1.5 mm}
\begin{align}
    \label{ladv}
    \min_{G_t}\max_{D_t}\mathcal{L}_{adv} &= \mathbb{E}_{x\sim p_{data}(x)}[\log D_t(x)] \\\notag &+
    \mathbb{E}_{z\sim p_{z}(z)}[\log{(1-D_t(G_t(z)))}],
    \vspace{-1.5 mm}
\end{align}
 where $z$ is a 1-D latent code sampled from noise distribution $p_z(z)$  (\eg, Gaussian), and $p_{data}(x)$ denotes the few-shot target data distribution. 
 Note that source data is inaccessible. 
 In fine-tuning, the weights of $G_s$ (and $D_{s}$) are used to initialize $G_t$ (and $D_{t}$). See Figure \ref{bad_knowledge_transfer}(a). The main goal of FSIG is to learn $G_t$ to capture $p_{data}(x)$.
%  , given $p_{data}(x)$ \textit{defined by few-shot target data}, is to estimate the true target distribution that (potentially) \textit{defined by abundant images}.

% Fine-tuning using Eqn. \ref{ladv} forms the basis of SOTA FSIG.
 To alleviate mode collapse due to very limited target samples,  recent methods {augment} fine-tuning with knowledge preservation to carefully select and preserve subset of source  knowledge during adaptation, \eg, freezing \cite{mo2020freezeD, yang2021one-shot-adaptation}, regularization \cite{li2020fig_EWC, ojha2021fig_cdc, zhao2022dcl} and modulation \cite{zhao2022fsig-ip} based methods. 
 The aim of these methods is to preserve knowledge that is deemed to be useful for target generator, \eg, improving the diversity of target sample generation \cite{ojha2021fig_cdc}. 
 For knowledge that is deemed to be less useful,  fine-tuning using Eqn.~(\ref{ladv}) is applied as a common practice to update such knowledge during adaptation.

% --------------------- End of Preliminary ----------------------- %

% ------------------------- Dissection -------------------------- %

\section{Incompatible Knowledge Transfer in FSIG}
\label{sec-3}
In this section, as our first contribution, we observe and identify the unnoticed issue of incompatible knowledge transfer in existing FSIG methods, 
and reveal that fine-tuning based knowledge update is inadequate to remove incompatible knowledge after adaptation.

% We remark that the issue of incompatible knowledge transfer has not been rigorously studied in previous works. In particular, previous works have primarily focused on setups where source and target domains are in close proximity \eg, {\small \texttt{Human Face} $\rightarrow$ \texttt{Face with Sunglasses}} \cite{ojha2021fig_cdc, zhao2022dcl}.
%  Under such setups most knowledge of source domain is compatible with the target domain, and consequently, the issue of incompatible knowledge is not significant and well noticed.
%  However, when source and target domains are dissimilar, \eg, 
%  {\small \texttt{Human Face} $\rightarrow$ \texttt{Cat Face}} in Figure~\ref{bad_knowledge_transfer}, we  observe that the incompatible knowledge transfer is severe.
 
 % originally new paragraph starts here
 To support our claim and %to discover 
 figure out the root cause of the observed incompatible knowledge transfer, 
 we apply GAN dissection \cite{bau2019gan, bau2017network}, a framework that can identify the correspondence between filters and the semantic segmentation of a particular object class (\eg, tree) across different images,
 to disclose %expose 
 filters that retain incompatible knowledge after fine-tuning.
 Overall, our main findings establish that existing SOTA methods fail to address this issue and uncover the root cause of incompatible knowledge transfer. 
%  In the next, we discuss the detailed analysis, experiment setups, and results.
 
 \begin{figure*}[t]
    \centering
    \includegraphics[width=\textwidth]{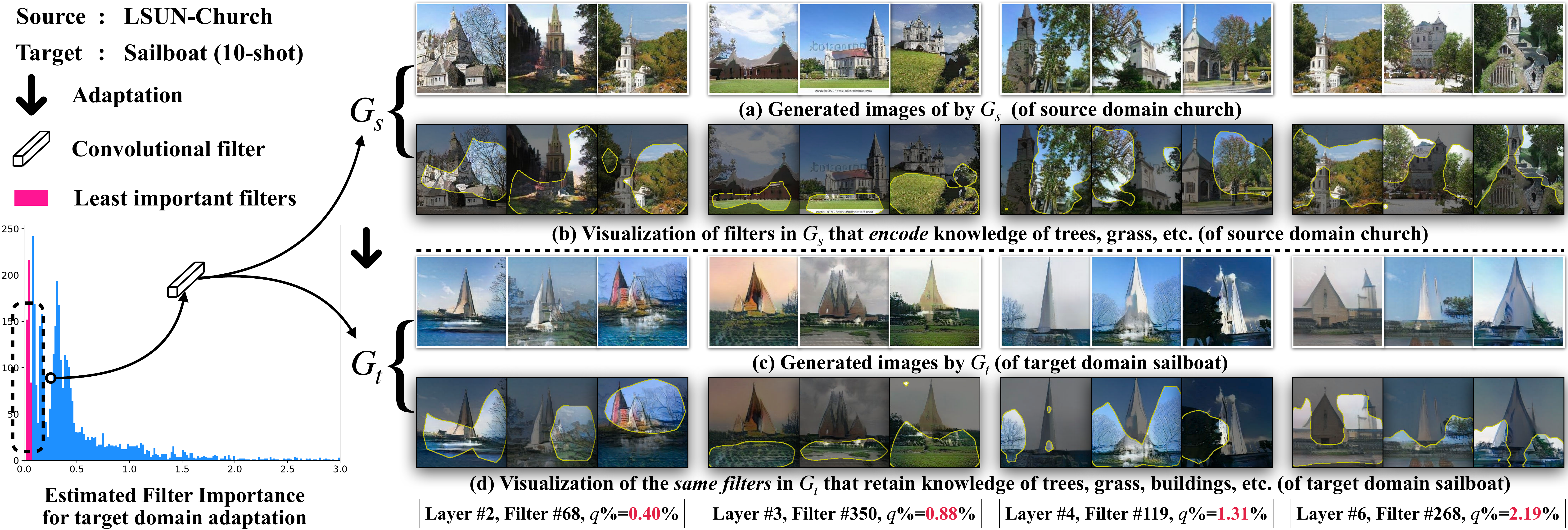}
    \vspace*{-5 mm}
    \caption{
    {\bf 
    Investigating the cause of incompatible knowledge transfer.}
    % In this example, we focus on 
    % {\small \texttt{Church} $\rightarrow$ \texttt{Sailboat}}.
    % To understand the root cause of incompatible knowledge transfer, 
    Since in Figure~\ref{bad_knowledge_transfer} we observe that {\em fine-tuning} commonly used in SOTA methods \cite{li2020fig_EWC,zhao2022fsig-ip} is inadequate for preventing incompatible knowledge transfer, we apply GAN dissection \cite{bau2019gan} to identify interpretable filters whose feature maps are highly correlated to the region of an object class (\eg, trees) across different images. We discover that the incompatible knowledge (\eg, trees, grass and buildings to Sailboat domain in this example) is correlated to the filters of $G_s$ that are deemed to be unimportant/irrelevant for target domain 
    (estimated via IP \cite{zhao2022fsig-ip}). 
    We use the quantile value $q\%$ to indicates the filter importance compared to all filters in $G_{s}$.
    % we apply the method in \cite{zhao2022fsig-ip} to estimate filters importance for target domain adaptation in $G_s$.
    % Some of these low importance filters (as determined by \cite{zhao2022fsig-ip}) encode knowledge of tree, building, grass as shown in the Figure.
    In \cite{zhao2022fsig-ip}, fine-tuning is applied during adaptation to update these low importance filters.
    % as they are determined to be irrelevant to the target domain
    % yunqing: Nov-11: see if there is smoother way to state
    Surprisingly, 
    % we apply GAN dissection \cite{bau2019gan} to visualize the knowledge {\em in the same filters} in $G_t$ .
    we observe that similar knowledge (tree, building, grass) remains in the \textit{same filters} in $G_t$ after fine-tuning.
    As this knowledge is incompatible to the target domain, it degrades the realisticness of synthetic images substantially. 
    Additional examples are in Supplement.
    %\textbf{Best viewed in color with zooming in.}
    }
    \label{fig_dissection}
    \vspace*{-1 mm}
\end{figure*}
 
 \subsection{Investigating incompatible knowledge}
 
 % motivation for our analysis
 Prior SOTA FSIG methods~\cite{li2020fig_EWC, ojha2021fig_cdc, zhao2022dcl, zhao2022fsig-ip} propose different knowledge preservation criteria to select pretrained source knowledge for few-shot adaptation. 
 The adaptation is typically done by fine-tuning the source generator (via 
%  $\mathcal{L}_{adv}$ in
 Eqn.~(\ref{ladv})) with the few-shot target samples.
 An assumption in these methods is that fine-tuning can adapt the source generator to the target one such that the irrelevant and incompatible source knowledge can be dropped or updated.
 
%  In this section, we focus on this question:
%  given that all recent SOTA FSIG methods use fine-tuning for adaptation, will incompatible knowledge for the target (\eg trees for the {\small\texttt{Sailboat}} domain) from $G_{s}$ be transferred to $G_{t}$? In other words, can fine-tuning based \textit{knowledge update} really remove incompatible knowledge for the target? 
 In this work, we show that the assumption becomes invalid in the cases where the source and target domains are semantically distant (\eg, {\small \texttt{Human Face} $\rightarrow$ \texttt{Cat Face}} in Figure~\ref{bad_knowledge_transfer}), where the incompatible knowledge transfer severely hurts the realisticness of the generated images.
%  We further explore how to \textit{detect} the transference of such incompatible knowledge.
 We note that this has not been well studied in prior SOTA FSIG works as they mainly focus on \textit{knowledge preservation} from the source (see Sec. \ref{sec-2}), while little attention has been paid to incompatible knowledge transfer with fine-tuning based \textit{knowledge update}.

 % connection between knowledge and filters
In convolutional neural networks, each filter 
%  $\mathbf{W} \in \mathbb{R}^{c^{in} \times k \times k}$ ($c^{in}$ is the number of input features, $k$ is the spatial size of the filter) 
can be viewed as an encoding of a specific part of knowledge \cite{bau2019gan, bau2017network}. Intuitively, in generative models, such knowledge could be either low-level textures (\eg, fur) or high-level human-interpretable concepts (\eg, eyes).  
 % why analyze filter?
Therefore, we hypothesize that clues for incompatible knowledge transfer could be found by attending to filters of the generator.
Recently, AdAM \cite{zhao2022fsig-ip} proposes an importance probing (IP) method to determine if a source GAN filter is important for adaptation 
%  Filters of high importance are preserved, and filters of low importance are fine-tuned.
%  They use Fisher Information (FI) \cite{mika1999fisher} as measurement for importance estimation.
and achieves impressive performance.
% in different FSIG setups.
% which implies the effectiveness of IP,
We adopt IP to evaluate the source generator filter importance for target domain adaptation in our analysis (we include a brief introduction of IP in Supplement).
 % we examine the knowledge encoded in low importance filters before and after fine-tuning using GAN dissection.
We propose two experiments at different granularity:

%  $\bullet$ 
 \textbf{Exp-1:} \textbf{Generate images with fixed generator input.}
 We visualize the generated images via different methods. To understand the knowledge transference before and after adaptation, we use the \textit{same noise} as input to source and target generator. Conceptually, this provides us an intuitive and direct comparison of knowledge transference.

 % see if there is a better name
%  $\bullet$ 
 \textbf{Exp-2:} \textbf{Dissect pretrained and adapted generator.}
%  To understand the relevancy between filters with different importance and generated images, 
%  we label ${G}_s$ filters with the estimated importance (via IP \cite{zhao2022fsig-ip}). 
%  After adaptation, we apply GAN dissection \cite{bau2019gan} and visualize the semantic features corresponding to the \textit{same filters} to $G_{s}$ and $G_{t}$. 
  To find the filters that are mostly correlated to a specific type of knowledge across different images (\eg, source features that are incompatible to the target) 
  and 
  track their transference before and after adaptation,
  we label ${G}_s$ filters with the estimated importance (via IP \cite{zhao2022fsig-ip}) 
  and 
  apply GAN dissection \cite{bau2019gan} to visualize the semantic features corresponding to the \textit{same filters} to $G_{s}$ and $G_{t}$. 
  
  These experiments could help us understand the knowledge transference before and after adaptation at both gross granularity (visualization of generated images in pixel space) and fine granularity (dissection of $G_{s}$ and $G_{t}$ in filter space).
  Next, we discuss the setups and results.

 \subsection{Experiment setups}
 \textbf{Model and dataset:} 
 In Exp-1, we use ProgressiveGAN (``ProGAN'') \cite{karras2017progan} and StyleGAN-V2 \cite{karras2020styleganv2} as GAN architectures. FFHQ \cite{karras2018styleGANv1}, LSUN-Church \cite{yu15lsun} are source domains; 10-shot AFHQ-Cat \cite{choi2020starganv2} and Sailboat are target domains. 
 Since Exp-2 is limited by the GAN architecture and segmentation model/dataset used in the original dissection work \cite{bau2019gan}, that are more suitable for scene-based scenarios \cite{yu15lsun}, we only use ProGAN with LSUN-Church \cite{yu15lsun} as the source domain and 10-shot Sailboat as target domain.
 Nevertheless, we remark that our analysis applies to a wide range of GAN architectures and domains.
 In all experiments, we use resolution 256 x 256 for adaptation.
 
 \textbf{Evaluation methods:}
 $G_{s}$ is the source generator.
 In Exp-1, we evaluate the baseline method, TGAN \cite{wang2018transferringGAN}, and recent SOTA, EWC \cite{li2020fig_EWC}, AdAM \cite{zhao2022fsig-ip}. 
 In Exp-2, we use AdAM \cite{zhao2022fsig-ip} for dissection (since we use IP to evaluate source filter importance for target adaptation). 
 We follow AdAM \cite{zhao2022fsig-ip} to use Fisher Information (FI) \cite{mika1999fisher} as importance measurement in IP.
 The dissection results of other methods, \eg, EWC \cite{li2020fig_EWC}, are in Supplement.
%  After adaptation, we use the source generator $G_s$ and the adapted generator $G_{t}$ for visualization and dissection analysis.
 
 \subsection{Results and analysis}
  We reveal that existing SOTA FSIG methods with a focus on source knowledge preservation lead to the transfer of incompatible knowledge. 
  More importantly, the \textit{root cause} of such incompatible knowledge transfer is the least important filters in $G_s$ determined to be irrelevant to the target domain adaptation, and fine-tuning is not adequate for removing the incompatible knowledge after adaptation.
%   Here ``incompatible knowledge'' denotes the semantic features from source that is no-good for the target. For example, trees, grass and building structures are transferred in {Church $\rightarrow$ Sailboat} setup, and hair style, spectacles and artifacts transferred in {FFHQ $\rightarrow$ AFHQ-Cat} setup, as Figure \ref{bad_knowledge_transfer}.
  Specifically, we summarize our observations in Figure \ref{bad_knowledge_transfer} and Figure \ref{fig_dissection}:
 
    {\textcolor{WildStrawberry}{\textbf{Observation 1:}}}
    In Figure \ref{bad_knowledge_transfer} \textbf{(c)}, we visualize the generated images by different methods with fixed noise input.
    % Interestingly, the incompatible features are indeed transferred after adaptation given different knowledge preservation criteria, \eg, ``tree on sea'' where ``tree'' is from Church domain and ``Cat with glasses'' where ``glasses'' is from FFHQ domain, and these incompatible source features severely curtails the realism of generated images after adaptation. 
    % We emphasize that the similar observation can be found on TGAN \cite{wang2018transferringGAN}, \ie, simple fine-tune method for knowledge update and no explicit knowledge preservation. 
    % In contrast, our method (to be discussed) can address this issue.
    Interestingly, features that are incompatible for the target domain are indeed transferred after adaptation with different knowledge preservation criteria, \eg, ``tree on sea'' where ``tree'' is from Church domain, and ``Cat with glasses'' where ``glasses'' is from FFHQ domain. All these incompatible source features severely curtail the realism of generated target images. 
    % We emphasize that the similar observation can be found on TGAN \cite{wang2018transferringGAN}, \ie, simple fine-tune method for knowledge update and no explicit knowledge preservation. 
    Similar observation can be made on TGAN \cite{wang2018transferringGAN}, \ie, simple fine-tuning based method without explicit knowledge preservation. 
    In contrast, our method (we discuss it in Sec. \ref{sec-4}) can address this issue.
     
    {\textcolor{WildStrawberry}{\textbf{Observation 2:}}}
    In Figure \ref{fig_dissection}, we dissect and visualize the incompatible features observed in Figure \ref{bad_knowledge_transfer}, and find their mostly correlated filters in $G_{s}$ and $G_{t}$.
    Surprisingly, we find that the filters in $G_{s}$ identified with \textit{least} importance for target domain
    % (we use quantile $q\%$ to indicate the ratio of filter importance over the largest importance value of all filters) 
    are \textit{mostly} relevant to incompatible features transferred from the source, which is the root cause of degradation of realism of generated images. 
    After adaptation, the same filters will still cause the same type of incompatible features, and  fine-tuning for knowledge update cannot effectively address this issue.
    When the target domains become distant, this observation is more obvious. 
    % We use AdAM for example in Figure \ref{fig_dissection}, and we have similar observation in other methods (\eg, EWC \cite{li2020fig_EWC}), which is included in Supplement.

 \begin{figure*}[t]
    \centering
    \includegraphics[width=\textwidth]{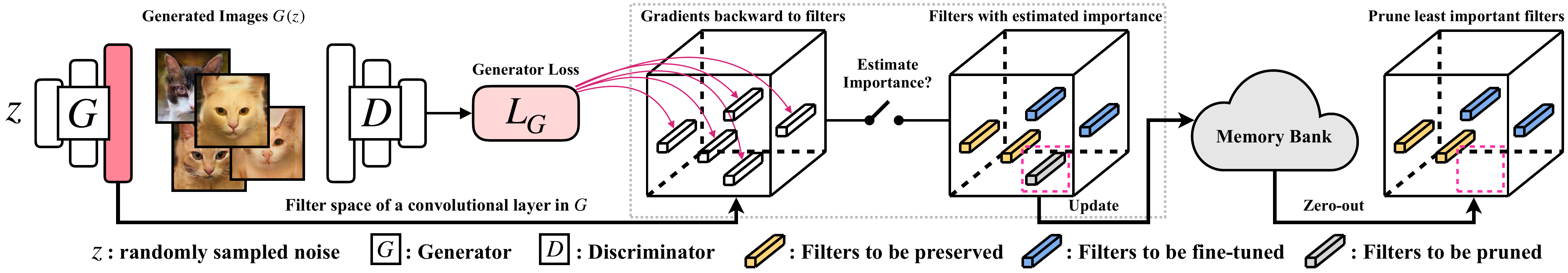}
    \vspace*{-7 mm}
    \caption{
    \textbf{Overview of the proposed method.}
    % In contrast to most prior works that focus only on \textit{source knowledge preservation}, 
    To {\bf R}emove {\bf I}n-{\bf C}ompatible {\bf K}nowledge during target adaptation,
    we propose \textit{knowledge truncation}, a novel concept for FSIG via pruning filters that are deemed with least importance for target domain adaptation.
    During training, we estimate the importance of filters for the target domain every certain iterations. 
    After that, we apply a fixed threshold to determine whether a filter should be pruned, and such decision will be maintained in a lightweight memory bank that is updated regularly upon importance estimation. 
    Similar to some prior works \cite{li2020fig_EWC, zhao2022dcl, zhao2022fsig-ip} that preserve useful source knowledge for adaptation, we preserve filters that are deemed to be important to the target domain by freezing them, and we fine-tune the rest of filters via Eqn.~(\ref{ladv}) for adaptation.
    % The operation to a specific filter in the next training iterations (\ie, before next importance evaluation) can be looked up in the memory bank.
    }
    \label{fig_method}
    \vspace*{-1 mm}
\end{figure*}

 Our analysis uncovers an important issue of existing SOTA FSIG methods: 
 source domain features that are incompatible to the target domain are transferred to the target generator, with various {knowledge preservation} criteria. The incompatible knowledge from the source is highly correlated to filters that are deemed to be irrelevant to the target domain, and fine-tuning based knowledge update cannot effectively address this issue.
%  This reveals the inadequacy of fine-tuning, which cannot effectively remove the features incompatible of the target. 
 This motivates us to remove the incompatible source features during FSIG adaptation.

\tocless

\section{Proposed Method}
\label{sec-4}

 Based on our analysis and observation in Sec. \ref{sec-3},
 we argue that {\bf R}emoving {\bf I}n-{\bf C}ompatible {\bf K}nowledge ({\bf RICK}) for the target domain is similarly important as preserving knowledge useful for the target domain to achieve improved quality of generated samples. 
 In contrast to most prior works that only propose different knowledge preservation criteria, \eg, freezing \cite{mo2020freezeD}, regularization \cite{li2020fig_EWC, ojha2021fig_cdc, zhao2022dcl} or modulation \cite{zhao2022fsig-ip},
 we propose knowledge truncation, a novel and complementary concept in existing methods for FSIG. 
 Overall, our proposed method, named RICK, is summarized in Figure \ref{fig_method}, and Algorithm {\color{red} 1} in Supplement. 
 
 \subsection{Knowledge truncation via network pruning}
 \label{sec-5.1}
 Pruning \cite{han2015learning, prune_redundancy, sui2021chip, hoefler2021sparsity} has been one of the useful tools to achieve a compact neural network with comparable performance to a larger, entire model.
 Early efforts on compacting networks focus on model acceleration \cite{he2019filter, he2020learning_acc}, inference efficiency \cite{li2020train}, and deployment \cite{lin2017runtime, wang2020apq}, which target at the discriminative tasks, \eg, image classification \cite{shen2022prune} and machine translation \cite{liang2021finding}, often by removing least important neurons (the definition of importance could be various and is discussed in Sec. \ref{sec-design}). 
 In contrast to prior works on network pruning that pursue the sparsity in the model, we aim to improve the quality of generated images via \textit{removing least important filters relevant to incompatible knowledge to the target domain}, particularly, in FSIG tasks. 
% {
% \color{red}
% yunqing: editing
% }

 Our proposed method contains two major steps: 
 1) a lightweight filter importance estimation on-the-fly during adaptation; 
 and 
 2) determine operations to filters based on their estimated importance.
 In step 1), we leverage the gradient information during adaptation to evaluate the filter importance for target adaptation every certain iterations. 
 Then in step 2), based on the estimated filter importance, we prune the filters with least importance, which are deemed to be irrelevant to the target domain
 to remove the incompatible knowledge for adaptation.
 Meanwhile, we preserve the filters with high importance to achieve knowledge preservation in FSIG, and fine-tune the rest of filters to let the source generator adapt to the target domain.
 
 \textbf{Proposed filter importance estimation.}
 We estimate the importance of each filter by leveraging the on-the-fly gradient information during FSIG adaptation. 
 We denote a filter as $\mathbf{W} \in \mathbb{R}^{c^{in} \times k \times k}$, where $k$ is the spatial size of the filter and $c^{in}$ is the dimension (number) of the input feature maps. 
 We use Fisher Information (FI) \cite{mika1999fisher} as importance estimator for each filter $\mathcal{F}(\mathbf{W})$
 (to be further discussed in Sec. \ref{sec-design}) 
 that could tell quantitative information of compatibility between filter weights and the FSIG task \cite{zhao2022fsig-ip, achille2019task2vec}:
\begin{equation}
    \label{eq:fisher_information}
    \vspace{-0mm}
    \mathcal{F}(\mathbf{W}) = \mathbb{E} \big[- \frac{\partial^2}{\partial\mathbf{W}^2} \mathcal{L}_{G}(x|\mathbf{W}) \big], 
    \vspace{-0mm}
    % \hspace{1 mm}
    % \text{if ($j$\%$\mathcal{T}$)=0},
\end{equation}
 where $\mathcal{L}_{G}$ is the binary cross-entropy loss computed with output from the discriminator. 
 $x$ denotes a set of generated images.
 In practice, we use first-order approximation of FI \cite{achille2019task2vec} to lower the computational cost.
 
 Our filter importance estimation for knowledge selection is lightweight and highly efficient: 
 compared to prior SOTA methods that propose different knowledge selection criteria (though they only focus on knowledge preservation), our method does not require external models to provide additional information during adaptation \cite{ojha2021fig_cdc, zhao2022dcl}, nor introduce additional learnable parameters and pre-adaptation iterations for importance estimation \cite{zhao2022fsig-ip}, and it takes benefits from the output of $G_{t}$ and $D_{t}$ during training.

 \textbf{Proposed knowledge truncation via filter pruning.}
 % idea
 In Sec. \ref{sec-3}, we have shown rich evidence that least important filters are relevant to semantic features incompatible to the target domain (\eg ``Tree on sea'' or ``Building structure on sea''). 
 Importantly, given different knowledge preservation criteria, fine-tuning based knowledge update cannot properly remove incompatible knowledge after adaptation. 
 Therefore, we propose a simple and novel method for knowledge truncation via pruning (zeroing-out) the filters with least importance for adaptation. 
 
 % ops of prune
 Specifically, after the estimation of filer importance in step 1), for the $i$-th filter $\mathbf{W}^{i}$ in the network, we apply a threshold ($q\%$, \ie, the quantile of its importance compared to all filters) to determine whether $\mathbf{W}^{i}$ should be pruned
%  {\color{red} eqn to be improved}
:
\begin{equation}
    \label{eq:prune}
    \mathbf{W}^{i} \gets \textbf{0}, 
    \hspace{1mm} \text{if}
    \hspace{1mm}
    \mathcal{F}(\mathbf{W}^{i}) \textless q\%
    % \hspace{1 mm}
    % \text{if ($j$\%$\mathcal{T}$)=0},
\end{equation}
 We remark that, once a filter is determined to be pruned, it will no longer be involved in training/inference and will not be recovered in the rest of training iterations. 
 The knowledge truncation is applied to both generator and discriminator, and we use separate thresholds to $G_{t}$ and $D_{t}$. 
 Since we regularly estimate the filter importance during adaptation and the ``non-recoverable'' attribute of pruned filters, the amount of zeroized filters using Eqn.~(\ref{eq:prune}) will accumulate to a specific value $p\%$, at the end of the adaptation.
 
 % mention knowledge preserve and update
 Similar to prior works that focus on knowledge preservation \cite{li2020fig_EWC, ojha2021fig_cdc, zhao2022dcl, zhao2022fsig-ip} and propose different knowledge selection criteria, we preserve the filters with high estimated importance for adaptation via freezing the filters during training. 
 For the rest of filters, we simply let them fine-tune using Eqn.~(\ref{ladv}). 
 Whether the filter needs to be fine-tuned or preserved depends dynamically on it's importance for target.
%  The filters that are fine-tuned could also be switched to be preserved if they are deemed to be important for target adaptation in the next importance estimation, and vice versa. 
 We discuss the effect of selecting high importance filters in Supplement. 
 Since we estimate the filter importance multiple times during adaptation, the operations to a specific filter may change after different evaluations, except the case that the filter is pruned and will not be recovered. 

%  We also dissect the generator and reveal that the high importance filters correspond to knowledge useful for the target domain.

%   {
%  \color{red}
%  editing
%  }

 \subsection{Design choice}
 \label{sec-design}
%  The importance of different filters (knowledge) could be \textit{dynamic} in different adaptation iterations. 
%  Different from post-evaluation in discriminative tasks and before-adaptation evaluation in \cite{zhao2022fsig-ip}, which may lead to additional training cost (in \cite{zhao2022fsig-ip} they need 500 training iterations for evaluation before main adaptation),  
 Here we discuss the design choice of our proposed method and adopted importance measurement. Since we dynamically evaluate the filter importance every certain iterations, we need to maintain the operation to each filter 
 (could be ``preserve'', ``fine-tune'' or ``prune'') until the next estimation.
 To lower the compute cost, we maintain the determination of operation to each filter (obtained via estimated filter importance) in a lightweight memory bank $\mathcal{M}$: for each high dimension filter $\mathbf{W} \in \mathbb{R}^{c^{in} \times k \times k}$, 
 we only need a single character to record the corresponding operation in $\mathcal{M}$. 
 For example, for StyleGAN-V2 \cite{karras2020styleganv2} used in main experiments whose generator contains $\sim$30M parameters 
 \footnote{\href{https://github.com/rosinality/stylegan2-pytorch}{\textcolor{RubineRed}{https://github.com/rosinality/stylegan2-pytorch}}}, $\mathcal{M}$ is a one dimension array with $\texttt{size}(\mathcal{M})\sim$ 5,000.
 
 % 2. importance measure
 Similar to prior works \cite{achille2019task2vec, li2020fig_EWC, zhao2022fsig-ip}, we use fisher information \cite{kirkpatrick2017ewc_pnas} as importance measurement to estimate how well a network parameter (filter in our work) does on the adaptation task \cite{achille2019task2vec}. 
 We note that there are alternative measurements to estimate the filter importance for adaptation, \eg, class salience \cite{simonyan2013class_salience} or reconstruction loss \cite{li2020fig_EWC}. 
 In Supplement, we conduct a study and empirically find that we can achieve similar performance as FI. 
 Moreover, in Sec. \ref{sec-6}, we surprisingly find that even without pruning (\ie, a filter can only be preserved or fine-tuned), our proposed method can still achieve competitive performance compared to SOTA methods, which implies the effectiveness of the proposed dynamic importance estimator.

\tocless

\section{Experiments}
\label{sec-5}

% %------------------------------------------

% \begin{algorithm}[t]
% \caption{Algorithm Description of FPGM}
% \label{alg:FPGM}
% \begin{algorithmic}[1]
% \STATE  \textbf{Input:} training data: $\mathbf{X}$.
% \STATE  \textbf{Given}:  pruning rate $P_{i}$
% \STATE  \textbf{Initialize}: model parameter $\mathbf{W} = \{\mathbf{W} ^{(i)}, 0\leq i \leq L\}$   
% \FOR{$epoch=1$; $epoch \leq epoch_{max}$; $epoch++$}
% 	\STATE Update the model parameter $\mathbf{W}$ based on $\mathbf{X}$
% 	\FOR{$i=1$; $i \leq L $; $i++$}          
% 		\STATE Find  $N_{i+1}P_i$ filters that satisfy Equation~\ref{eq:7}
% 		\STATE Zeroize selected filters
% 	\ENDFOR
% \ENDFOR
% \STATE Obtain the compact model $\mathbf{W} ^{*}$ from $\mathbf{W}$
% \STATE \textbf{Output:} The compact model and its parameters $\mathbf{W} ^{*}$
% \end{algorithmic} 
% \end{algorithm}

% %-------------------------------------------------------

\begin{figure*}[t]
    \centering
    \includegraphics[width=\textwidth]{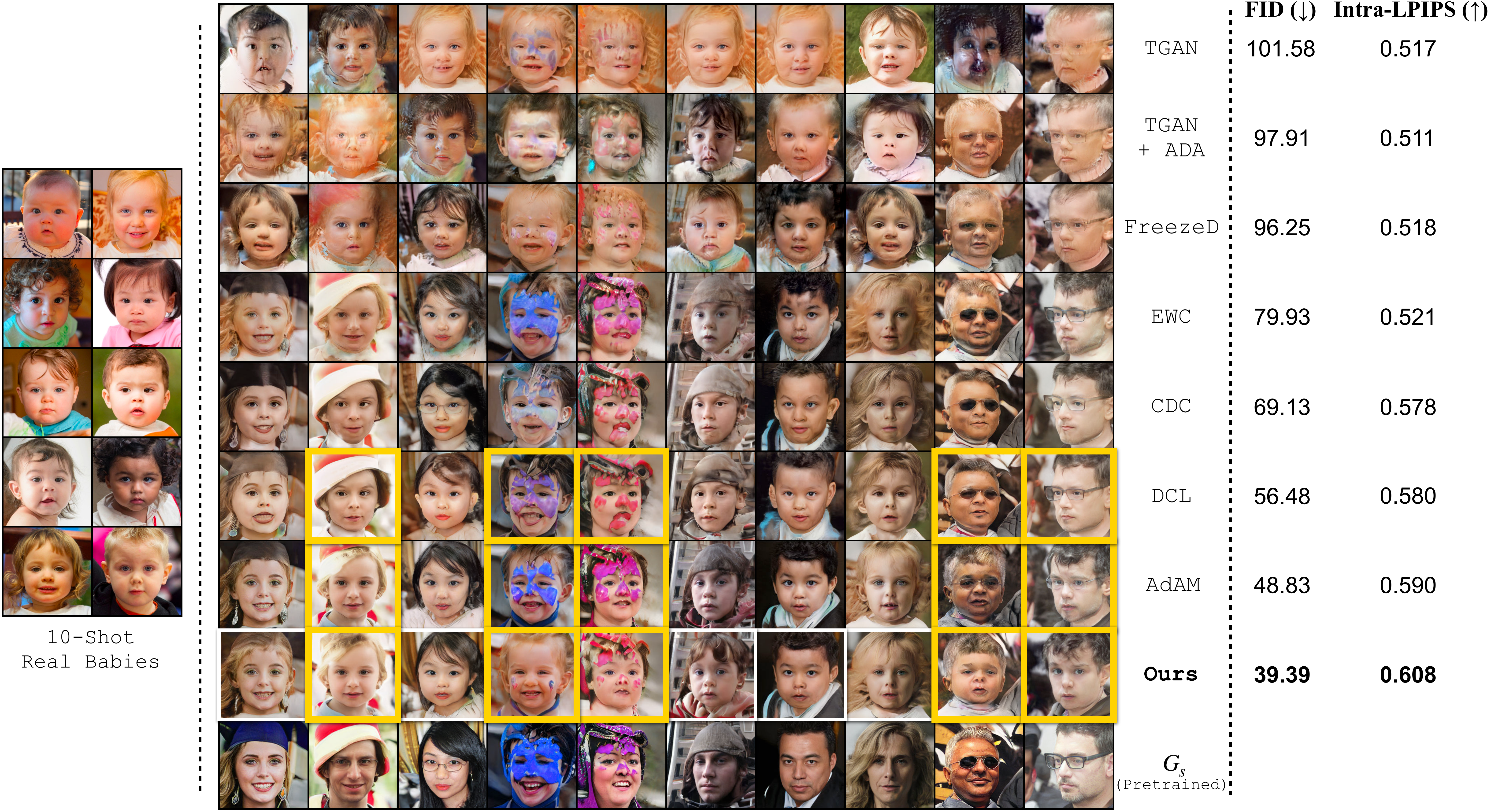}
    % \vspace{.2cm}
    \includegraphics[width=\textwidth]{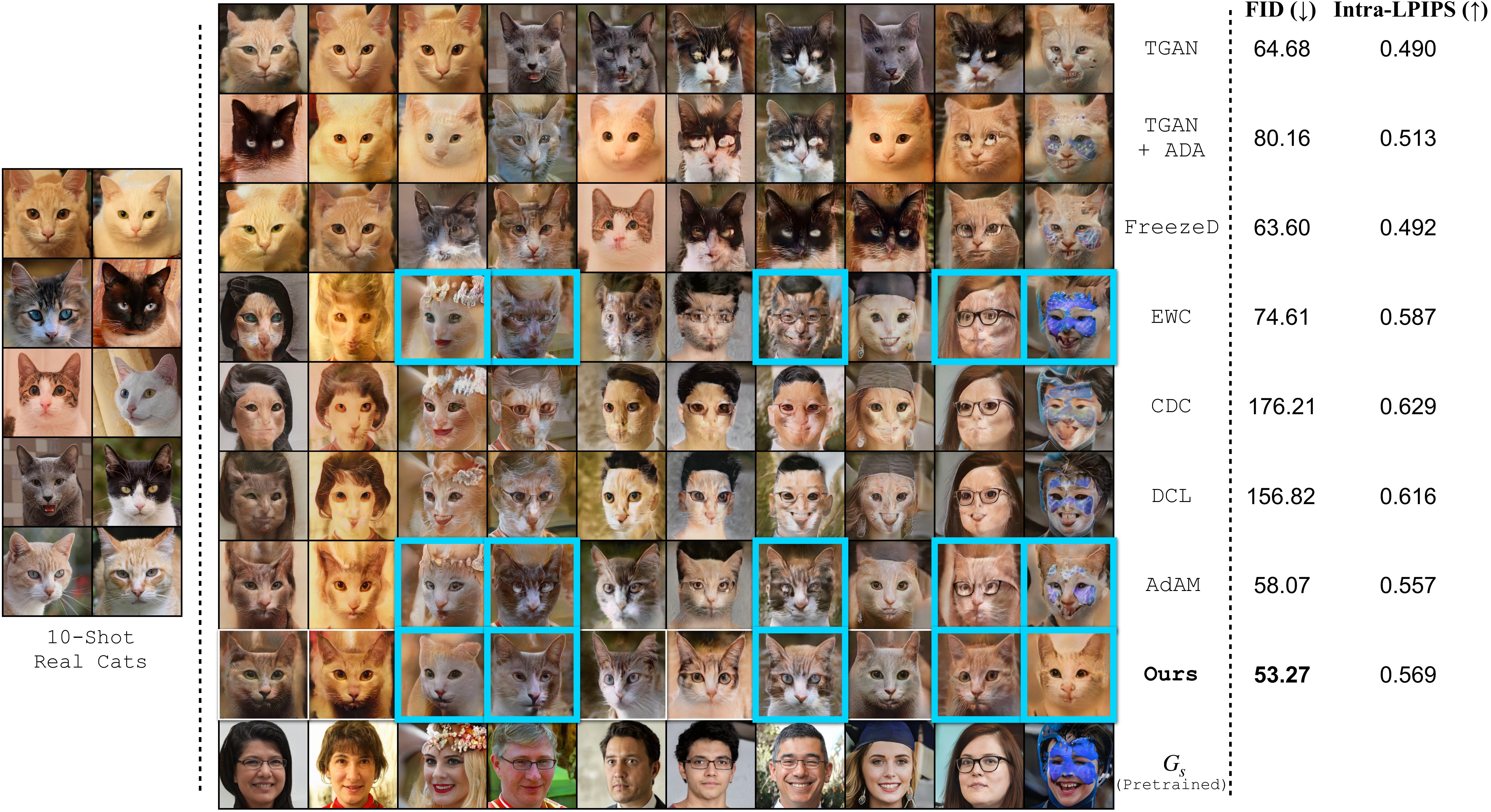}
    \caption{
    Experiment results of FSIG on few-shot target samples. FFHQ \cite{karras2018styleGANv1} is the source domain. 
    \textbf{Left}:
    10-shot real target samples for adaptation.
    \textbf{Mid}: We visualize the generated images using the adapted generator $G_t$ with different methods. Images in each column are from the same noise input. 
    It is noticeable that our method while preserving the source knowledge useful for the target domain, 
    reliably removes the incompatible source knowledge. 
    For example, in SOTA methods,
    \textbf{(Top {\color{Dandelion} \textbf{orange}}} frames) hat, doodle, sunglasses and beard are transferred and lead to generated babies with degraded realism; 
    \textbf{(Bottom {\color{cyan} \textbf{blue}}} frames) hat, glasses, human face texture and artifacts are transferred and lead to generated cats with low realism. 
    In contrast, our method can address this issue in different setups.
    \textbf{Right}:
    Quantitatively, we measure the quality and diversity of generated images via FID ($\downarrow$) \cite{heusel2017FID} and intra-LPIPS ($\uparrow$) \cite{ojha2021fig_cdc}. See details in Sec. \ref{sec-5}.  
    }
    \label{fig_comparison}
\end{figure*}

\begin{table*}[t]  
    \renewcommand{\arraystretch}{0.95}
    \centering
    \caption{
    We report FID ($\downarrow$) as quantitative results for FSIG (10-shot), FFHQ is the source domain. 
    We compare our proposed with other baseline and SOTA methods over six target datasets, including the challenging setups that target domains are distant to the source (\eg. AFHQ datasets). 
    % We randomly generate 5,000 images using the adapted generator and compare with the entire target dataset.
    We emphasize that, for SOTA methods that focus only on knowledge preservation (\eg, EWC \cite{li2020fig_EWC}, CDC \cite{ojha2021fig_cdc}, DCL \cite{zhao2022dcl}, AdAM \cite{zhao2022fsig-ip}),  
    incompatible source knowledge is still transferred and therefore it curtails the quality of generated images. 
    In contrast, our methods can remove the knowledge incompatible for the target and preserve the knowledge important for the target, therefore achieve improved quality of generated images. 
    % We additionally include more results, \eg, KID ($\downarrow$) \cite{binkowski2018kid_score} in Supplement.
    }
    \vspace{-3 mm}
    \begin{adjustbox}{width=\textwidth,center}
        \begin{tabular}{l| l l l l l l l l}
        \toprule
        \textbf{Target Domain}
         & \textbf{Babies} \cite{ojha2021fig_cdc}
         & \textbf{Sunglasses} \cite{ojha2021fig_cdc}
         & \textbf{MetFaces} \cite{karras2020ADA}
        %  & \textbf{AFHQ-Cat} \cite{choi2020starganv2}
        & \textbf{AFHQ-Cat} \cite{choi2020starganv2}
        %  & \textbf{AFHQ-Dog} \cite{choi2020starganv2}
        & \textbf{AFHQ-Dog} \cite{choi2020starganv2}
        %  & \textbf{AFHQ-Wild} \cite{choi2020starganv2}
        & \textbf{AFHQ-Wild} \cite{choi2020starganv2}
         \\ 
        \hline
        TGAN \cite{wang2018transferringGAN} & $101.58$ & $55.97$ & $76.81$ & $64.68$ & $151.46$ & $81.30$    \\ \midrule
        TGAN+ADA \cite{karras2020ADA} & $97.91$ & $53.64$ & $75.82$ & $80.16$ & $162.63$ & $81.55$ \\ \midrule
        FreezeD \cite{mo2020freezeD} & $96.25$ & $46.95$ & $73.33$ & $63.60$ & $157.98$ & $77.18$ \\ \midrule
        CDC \cite{ojha2021fig_cdc} & $69.13$ & $41.45$ & $65.45$ & $176.21$ & $170.95$ & $135.13$  \\ \midrule
        DCL \cite{zhao2022dcl} & $56.48$ & $37.66$ & $62.35$ & $156.82$ & $171.42$ & $115.93$ \\ 
        \midrule
        EWC \cite{li2020fig_EWC} & $79.93$ & $49.41$ & $62.67$ & $74.61$ & $158.78$ & $92.83$   \\ 
        EWC + RICK (\textbf{Ours}) 
        & 
        {$68.22({\color{RoyalBlue} -\mathbf{11.71}})$}
        & 
        {$39.53({\color{RoyalBlue} -\mathbf{9.88}})$}
        & 
        {$54.7({\color{RoyalBlue} -\mathbf{7.97}})$}
        & 
        {$64.35({\color{RoyalBlue} -\mathbf{10.26}})$}
        & 
        {$124.50({\color{RoyalBlue} -\mathbf{34.28}})$}
        & 
        {$56.83({\color{RoyalBlue} -\mathbf{36.00}})$}
        \\
        \midrule
        AdAM \cite{zhao2022fsig-ip} & {$48.83$} & {$28.03$} & {$51.34$} & {$58.07$} & {$100.91$} & {$36.87$} \\
        AdAM + RICK (\textbf{Ours}) & {$43.12({\color{RoyalBlue} -\mathbf{5.71}})$} & 
        {$26.25({\color{RoyalBlue} -\mathbf{1.78}})$}
        &
        {$49.47({\color{RoyalBlue} -\mathbf{1.87}})$} 
        & 
        {$53.94({\color{RoyalBlue} -\mathbf{4.13}})$} 
        &
        {$100.35({\color{RoyalBlue} -\mathbf{0.56}})$} 
        &
        {$35.54({\color{RoyalBlue} -\mathbf{1.33}})$} 
        \\
        \midrule
        \textbf{Ours} & 
        % \bm{$39.39$} & & & \bm{$53.27$}
        \bm{$39.39$} & \bm{$25.22$} & \bm{$48.53$} & \bm{$53.27$} & \bm{$98.71$} & \bm{$33.02$}  
        \\
        \bottomrule
        \end{tabular}
    \end{adjustbox}
    \vspace{-4 mm}
\label{table:fid}
\end{table*}

%\subsection{Implementation details.}
 % We discuss the main experiments and results in this section. 
 % {\color{RubineRed}{Code}} and additional results are in Supplement.

\textbf{Basic setups.} 
 For fair comparison, we strictly follow the experiment setups as previous works \cite{li2020fig_EWC, ojha2021fig_cdc, zhao2022dcl}, \eg in the choice of source and target domains and few-shot target samples. We employ StyleGAN-V2 \cite{karras2020styleganv2} 
 as the GAN architecture for pretraining and adaptation in main experiments, similar to previous works \cite{ojha2021fig_cdc, yang2021one-shot-adaptation, zhao2022dcl, xiao2022few, zhao2022fsig-ip}. 
 We train our models with on an NVIDIA A100 PCIe 40GB. We include more implementation details in Supplement.

 \textbf{Datasets and baseline methods.}
 We use the GAN pretrained on FFHQ \cite{karras2018styleGANv1} and the target domains that have different proximity to the source dataset: semantically related domains include Babies \cite{ojha2021fig_cdc}, Sunglasses \cite{ojha2021fig_cdc} and MetFaces \cite{karras2020ADA} 
 (we note that this is the common setup in prior works for FSIG); 
 distant domains include AFHQ-Cat, AFHQ-Dog and AFHQ-Wild \cite{choi2020starganv2} 
 (we note that this setup is more challenging compared to most prior works \cite{li2020fig_EWC, ojha2021fig_cdc, zhao2022dcl}).
 Adaptation on more target domains are in Supplement for comprehensive analysis. We use images with resolution 256 x 256 for adaptation. 
 Baseline methods used for comparison are introduced in Sec. \ref{sec-2}. 
 We additionally compare with ADA \cite{karras2020ADA}, a popular baseline approach that applies data augmentation during adaptation.

\subsection{Performance evaluation and comparison}
\textbf{Evaluation measurements.}
 We adopt three types of evaluation methods in the main paper. The popular metric Fréchet Inception Distance (FID $\downarrow$) \cite{heusel2017FID}
 %\footnote{\href{https://github.com/mseitzer/pytorch-fid}{\textcolor{RubineRed}{https://github.com/mseitzer/pytorch-fid}}}
 measures the distance between the fitted Gaussian distribution of the real and generated data. We also evaluate the diversity of the generated images using intra-LPIPS ($\uparrow$) \cite{ojha2021fig_cdc},
%  \footnote{\href{https://github.com/utkarshojha/few-shot-gan-adaptation}{\textcolor{RubineRed}{https://github.com/utkarshojha/few-shot-gan-adaptation}}}
 the perceptual distance of generated images to few-shot target samples (we include the pseudo-code in Supplement). Finally, we visualize the generated images and compare with different methods, often with a fixed noise input for fair comparison.

\textbf{Qualitative results.}
 In Figure \ref{fig_comparison}, we visualize the generated images by different methods before and after adaptation for comparison. 
 In each column, images are generated from the same noise input.
 We use FFHQ \cite{karras2018styleGANv1} as the source domain. Babies \cite{ojha2021fig_cdc} and AFHQ-Cat \cite{choi2020starganv2} are target domains with different semantic proximity to the source. We show that, our proposed method, while preserving useful source knowledge, reliably removes the incompatible knowledge to the target, and therefore achieves improved quality of generated images. 
 More results are in Supplement.
%  Since our method can estimate the filter importance dynamically during adaptation, it can \textit{preserve} important filters for the target via \texttt{stop\_gradient} operations and \textit{drop} filters that curtail the FSIG adaptation via \texttt{pruning}.
%  Therefore, our method can preserve meaningful knowledge from the source, and more importantly, it reliably removes the adversarial features from the source, \eg, (\textbf{Top}) sunglasses, hat and artifacts from the source and (\textbf{Bottom}) glasses, hair style and artifacts from the source, compared to all other baseline and SOTA methods.

\textbf{Quantitative results.}
 Considering that the whole target dataset often contains about 5,000 images (\eg, AFHQ-Cat \cite{choi2020starganv2}), following prior works \cite{ojha2021fig_cdc, zhao2022dcl, zhao2022fsig-ip} we randomly generate 5,000 images using the adapted generator and compare with the entire target dataset to compute FID ($\downarrow$). In Table \ref{table:fid}, we show complete FID results over six benchmark datasets. In Figure \ref{fig_comparison}, we also compute intra-LPIPS ($\uparrow$) as diversity measurement over 10-shot target samples and we report the FID using the same checkpoint. All these results show the effectiveness of our proposed method. 
%  Additional quantitative results including KID ($\downarrow$) \cite{binkowski2018kid_score} are in Supplement.

\tocless

 \begin{figure}[t]
    \centering
    \includegraphics[width=\columnwidth]{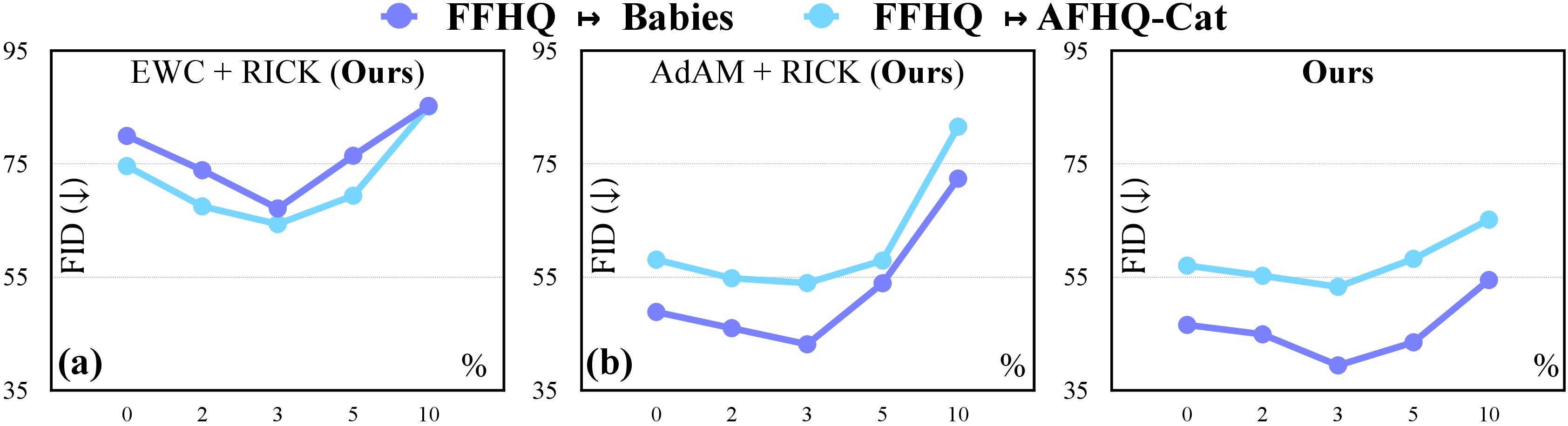}
    \vspace{-6 mm}
    \caption{
    FID ($\downarrow$) by pruning different percent (\%) of filters, which importance are estimated in different ways. 
    We note that prune 0\% filters in (\textbf{a}-\textbf{b}) indicates the original EWC \cite{li2020fig_EWC} and AdAM \cite{zhao2022fsig-ip}.
    % Ideally, if the amount of pruned filters exceeds a specific threshold, it may also degrade the quality of generated images. 
    % Based on the above observation, we apply a fixed threshold and prune 3\% least important filters consistently through different setups. 
    }
    \label{fig_prune_percent}
    \vspace{-5 mm}
\end{figure}

\subsection{Discussion}
\label{sec-6}
\textbf{Knowledge truncation with different methods.}
Ideally, our proposed concept of knowledge truncation for FSIG can be applied to different methods, as long as we can estimate the parameter importance (\eg, filter importance in our method). 
In literature, EWC \cite{li2020fig_EWC} and AdAM \cite{zhao2022fsig-ip} proposed different methods to evaluate the parameter importance: EWC directly estimates the parameter importance on the source dataset of $G_{s}$, while AdAM uses a modulation based method to estimate the $G_{s}$ parameter importance on the target dataset. Therefore, in Table \ref{table:fid}, we also show the results of applying our proposed knowledge truncation to EWC and AdAM. 
As our method can effectively remove the incompatible knowledge by pruning least important filters, we can achieve consistent improved performance on different datasets. 
More details are in Supplement.

\textbf{Prune different percent of filters.}
 We empirically study the impact of pruning different percent of filters. According to the results in Figure \ref{fig_prune_percent}, we prune different amounts of filters on three different methods. Ideally, if we prune more filters, some important knowledge will be removed and the performance will be degraded accordingly.
 Therefore, we prune 3\% (\ie, $p$=3 in Sec. \ref{sec-5.1}) filters in different setups, which can achieve considerable and stable improvements.

\textbf{Can we train longer to remove the incompatible knowledge?}
 Ideally, an intuitive and potentially useful way to remove the incompatible knowledge is to simply train longer iterations. 
 In Supplement, however, we conduct a study and show that for existing FSIG methods, as the target set contains only 10-shot training images, training longer iterations will let the generator overfit and tend to replicate the few-shot target samples, such that it can fool the discriminator. The diversity of generated images is degraded significantly.
 Therefore, it is important to remove incompatible knowledge before the overfitting becomes severe.

% There could be other general ablation studies, analysis, can include in Supplement. E.g., number of shots, adaptation to more domains.
% }

% \textbf{Ablation study of proposed method.} We show the effect of the proposed knowledge truncation and the dynamic evaluation on the performance in Table \ref{table:fid_ablation}.

% \textbf{Impact of filter importance threshold.}
%  In practice, the two thresholds, $t_{h}$ and $t_{l}$, control the amount of knowledge we preserve and remove.

% % \renewcommand{\arraystretch}{0.96}
% \begin{table}[t]  
%     \centering
%     \caption{
%     }
%     \begin{adjustbox}{width=0.95\columnwidth,center}
%         \begin{tabular}{l| c c c }
%         \toprule
%         \textbf{Target Domain}
%         & \textbf{Babies} \cite{ojha2021fig_cdc}
%         & \textbf{AFHQ-Cat} \cite{choi2020starganv2}
%          \\ 
%         \hline
%         TGAN \cite{wang2018transferringGAN} & $101.58$ & $64.68$ \\ \midrule
%         EWC \cite{li2020fig_EWC} & $79.93$ & $74.61$  \\
%         \midrule
%         AdaM \cite{zhao2022fsig-ip} & {$48.83$} & {$58.07$}    \\\midrule
%         AdAM + Prune 1\% (static)  & $40.70$ & $55.70$ \\
%         AdAM + Prune 1\% (dynamic) & \bm{$39.39$} & - \\\midrule
%         AdAM + Prune 3\% (static)  & $43.12$  & $53.94$  \\
%         AdAM + Prune 3\% (dynamic) & $41.05$ & \bm{$53.27$} \\
%         \bottomrule
%         \end{tabular}
%     \end{adjustbox}
% \label{table:fid_ablation}
% \end{table}

% Conclusion
\tocless

\section{Conclusion}
\label{sec-7}

 We tackle few-shot image generation (FSIG) in this work.  
 As the first contribution, we uncover the unnoticed issue of incompatible knowledge transfer of existing SOTA methods, which leads to significant degradation of generated image realism. 
 Surprisingly, we discover that the root cause of such incompatible knowledge transfer is the filters that are deemed to have least importance for target adaptation, and fine-tuning based SOTA methods cannot properly address this issue. 
 We therefore propose a novel concept, \textit{knowledge truncation} for FSIG, which aims to eradicate the incompatible knowledge via pruning  filters with least importance for adaptation. 
 Our proposed filter importance estimation takes benefits of the gradient information from dynamic training process, and it is lightweight on computational cost. 
 Through extensive experiments, we show that our proposed method can be applied to a wide range of adaptation setups with different GAN architectures. We achieve new state-of-the-art performance, including visually pleasant generated images without much incompatible knowledge transferred, and improved quantitative results.

{\bf Limitation and ethical concerns.}
%  We follow  previous work in experiment setups.
 The scale of our experiments is comparable to prior works. 
 Nevertheless, extensions of our 
 knowledge truncation method, additional datasets and generative models beyond GANs 
 (\eg, Variational Auto-Encoders \cite{van2017vq_vae} or Diffusion Models \cite{rombach2021highresolution})
 can be considered as future work.
 Our proposed FSIG methods may have negative societal impacts if it is used by malicious users.
 However, our work contributes to increased awareness about image generation with limited data.

%  In Supplement, we also discuss the broader impact, limitations, and future works of our research.

\section*{Acknowledgment}
% \textbf{Acknowledgment}. 
 {
 This research is supported by the National Research Foundation, Singapore under its AI Singapore Programmes (AISG Award No.: AISG2-RP-2021-021; AISG Award No.: AISG-100E2018-005). 
 This project is also supported by SUTD project PIE-SGP-AI-2018-01. 
 We thank anonymous reviewers for their insightful comments.
 
 }

% -------------------------------------------------------------------- %
% ----------------------------- End Sign ----------------------------- %

%%%%%%%%% REFERENCES
{\small
\bibliographystyle{ieee_fullname}
\bibliography{egbib}

\begin{thebibliography}{10}\itemsep=-1pt

\bibitem{achille2019task2vec}
Alessandro Achille, Michael Lam, Rahul Tewari, Avinash Ravichandran, Subhransu
  Maji, Charless~C Fowlkes, Stefano Soatto, and Pietro Perona.
\newblock Task2vec: Task embedding for meta-learning.
\newblock In {\em Proceedings of the IEEE/CVF International Conference on
  Computer Vision}, pages 6430--6439, 2019.

\bibitem{Afrasiyabi_2022_CVPR_few_shot_transfer}
Arman Afrasiyabi, Hugo Larochelle, Jean-Fran\c{c}ois Lalonde, and Christian
  Gagn\'e.
\newblock Matching feature sets for few-shot image classification.
\newblock In {\em Proceedings of the IEEE/CVF Conference on Computer Vision and
  Pattern Recognition (CVPR)}, pages 9014--9024, June 2022.

\bibitem{bau2017network}
David Bau, Bolei Zhou, Aditya Khosla, Aude Oliva, and Antonio Torralba.
\newblock Network dissection: Quantifying interpretability of deep visual
  representations.
\newblock In {\em Proceedings of the IEEE conference on computer vision and
  pattern recognition}, pages 6541--6549, 2017.

\bibitem{bau2019gan}
David Bau, Jun-Yan Zhu, Hendrik Strobelt, Bolei Zhou, Joshua~B Tenenbaum,
  William~T Freeman, and Antonio Torralba.
\newblock Gan dissection: Visualizing and understanding generative adversarial
  networks.
\newblock In {\em Proceedings of the International Conference on Learning
  Representations (ICLR)}, 2019.

\bibitem{binkowski2018kid_score}
Mikolaj Binkowski, Dougal~J. Sutherland, Michael Arbel, and Arthur Gretton.
\newblock Demystifying {MMD} {GAN}s.
\newblock In {\em International Conference on Learning Representations}, 2018.

\bibitem{brock2018bigGAN}
Andrew Brock, Jeff Donahue, and Karen Simonyan.
\newblock Large scale gan training for high fidelity natural image synthesis.
\newblock {\em International Conference on Learning Representations}, 2018.

\bibitem{chai2021ensembling}
Lucy Chai, Jun-Yan Zhu, Eli Shechtman, Phillip Isola, and Richard Zhang.
\newblock Ensembling with deep generative views.
\newblock In {\em CVPR}, 2021.

\bibitem{keshik2022ls-kd}
Keshigeyan Chandrasegaran, Ngoc-Trung Tran, Yunqing Zhao, and Ngai-Man Cheung.
\newblock Revisiting label smoothing and knowledge distillation compatibility:
  What was missing?
\newblock In {\em International Conference on Machine Learning}, pages
  2890--2916. PMLR, 2022.

\bibitem{choi2020starganv2}
Yunjey Choi, Youngjung Uh, Jaejun Yoo, and Jung-Woo Ha.
\newblock Stargan v2: Diverse image synthesis for multiple domains.
\newblock In {\em Proceedings of the IEEE Conference on Computer Vision and
  Pattern Recognition}, 2020.

\bibitem{cong2020gan_memory}
Yulai Cong, Miaoyun Zhao, Jianqiao Li, Sijia Wang, and Lawrence Carin.
\newblock Gan memory with no forgetting.
\newblock {\em Advances in Neural Information Processing Systems},
  33:16481--16494, 2020.

\bibitem{crockett1992face_otto_dix}
Dennis Crockett.
\newblock The most famous painting of the “golden twenties”? otto dix and
  the trench affair.
\newblock {\em Art Journal}, 51(1):72--80, 1992.

\bibitem{deng2009imagenet}
Jia Deng, Wei Dong, Richard Socher, Li-Jia Li, Kai Li, and Li Fei-Fei.
\newblock Imagenet: A large-scale hierarchical image database.
\newblock In {\em 2009 IEEE conference on computer vision and pattern
  recognition}, pages 248--255, 2009.

\bibitem{Evci:head2toe:icml2022}
Utku Evci, Vincent Dumoulin, Hugo Larochelle, and Michael~C. Mozer.
\newblock Head2toe: Utilizing intermediate representations for better transfer
  learning.
\newblock In {\em ICML}, 2022.

\bibitem{adar:liver:2018}
Maayan Frid-Adar, Idit Diamant, Eyal Klang, Michal Amitai, Jacob Goldberger,
  and Hayit Greenspan.
\newblock Gan-based synthetic medical image augmentation for increased cnn
  performance in liver lesion classification.
\newblock {\em Neurocomputing}, 321:321--331, 2018.

\bibitem{gong2022meta}
Jia Gong, Zhipeng Fan, Qiuhong Ke, Hossein Rahmani, and Jun Liu.
\newblock Meta agent teaming active learning for pose estimation.
\newblock In {\em Proceedings of the IEEE/CVF Conference on Computer Vision and
  Pattern Recognition}, pages 11079--11089, 2022.

\bibitem{gong2022diffpose}
Jia Gong, Foo Lin~Geng, Zhipeng Fan, and Jun Liu.
\newblock Diffpose: Toward more reliable 3d pose estimation.
\newblock In {\em Proceedings of the IEEE/CVF Conference on Computer Vision and
  Pattern Recognition}, 2023.

\bibitem{goodfellow2014GAN}
Ian Goodfellow, Jean Pouget-Abadie, Mehdi Mirza, Bing Xu, David Warde-Farley,
  Sherjil Ozair, Aaron Courville, and Yoshua Bengio.
\newblock Generative adversarial nets.
\newblock {\em Advances in neural information processing systems}, 27, 2014.

\bibitem{Gupta_2022_CVPR_detection_transfer}
Akshita Gupta, Sanath Narayan, K~J Joseph, Salman Khan, Fahad~Shahbaz Khan, and
  Mubarak Shah.
\newblock Ow-detr: Open-world detection transformer.
\newblock In {\em Proceedings of the IEEE/CVF Conference on Computer Vision and
  Pattern Recognition (CVPR)}, pages 9235--9244, June 2022.

\bibitem{han2015learning}
Song Han, Jeff Pool, John Tran, and William Dally.
\newblock Learning both weights and connections for efficient neural network.
\newblock {\em Advances in neural information processing systems}, 28, 2015.

\bibitem{he2020moco}
Kaiming He, Haoqi Fan, Yuxin Wu, Saining Xie, and Ross Girshick.
\newblock Momentum contrast for unsupervised visual representation learning.
\newblock In {\em Proceedings of the IEEE/CVF Conference on Computer Vision and
  Pattern Recognition}, pages 9729--9738, 2020.

\bibitem{he2020learning_acc}
Yang He, Yuhang Ding, Ping Liu, Linchao Zhu, Hanwang Zhang, and Yi Yang.
\newblock Learning filter pruning criteria for deep convolutional neural
  networks acceleration.
\newblock In {\em Proceedings of the IEEE/CVF conference on computer vision and
  pattern recognition}, pages 2009--2018, 2020.

\bibitem{he2019filter}
Yang He, Ping Liu, Ziwei Wang, Zhilan Hu, and Yi Yang.
\newblock Filter pruning via geometric median for deep convolutional neural
  networks acceleration.
\newblock In {\em Proceedings of the IEEE/CVF conference on computer vision and
  pattern recognition}, pages 4340--4349, 2019.

\bibitem{heusel2017FID}
Martin Heusel, Hubert Ramsauer, Thomas Unterthiner, Bernhard Nessler, and Sepp
  Hochreiter.
\newblock Gans trained by a two time-scale update rule converge to a local nash
  equilibrium.
\newblock {\em Advances in neural information processing systems}, 30, 2017.

\bibitem{ho2020denoising}
Jonathan Ho, Ajay Jain, and Pieter Abbeel.
\newblock Denoising diffusion probabilistic models.
\newblock {\em Advances in Neural Information Processing Systems},
  33:6840--6851, 2020.

\bibitem{hoefler2021sparsity}
Torsten Hoefler, Dan Alistarh, Tal Ben-Nun, Nikoli Dryden, and Alexandra Peste.
\newblock Sparsity in deep learning: Pruning and growth for efficient inference
  and training in neural networks.
\newblock {\em J. Mach. Learn. Res.}, 22(241):1--124, 2021.

\bibitem{Hu_2022_CVPR_few_shot_transfer}
Shell~Xu Hu, Da Li, Jan St\"uhmer, Minyoung Kim, and Timothy~M. Hospedales.
\newblock Pushing the limits of simple pipelines for few-shot learning:
  External data and fine-tuning make a difference.
\newblock In {\em Proceedings of the IEEE/CVF Conference on Computer Vision and
  Pattern Recognition (CVPR)}, pages 9068--9077, June 2022.

\bibitem{huang2022masked}
Jiaxing Huang, Kaiwen Cui, Dayan Guan, Aoran Xiao, Fangneng Zhan, Shijian Lu,
  Shengcai Liao, and Eric Xing.
\newblock Masked generative adversarial networks are data-efficient generation
  learners.
\newblock In Alice~H. Oh, Alekh Agarwal, Danielle Belgrave, and Kyunghyun Cho,
  editors, {\em Advances in Neural Information Processing Systems}, 2022.

\bibitem{karras2017progan}
Tero Karras, Timo Aila, Samuli Laine, and Jaakko Lehtinen.
\newblock Progressive growing of gans for improved quality, stability, and
  variation.
\newblock {\em arXiv preprint arXiv:1710.10196}, 2017.

\bibitem{karras2020ADA}
Tero Karras, Miika Aittala, Janne Hellsten, Samuli Laine, Jaakko Lehtinen, and
  Timo Aila.
\newblock Training generative adversarial networks with limited data.
\newblock {\em Advances in Neural Information Processing Systems},
  33:12104--12114, 2020.

\bibitem{Karras2021styleganv3}
Tero Karras, Miika Aittala, Samuli Laine, Erik H\"ark\"onen, Janne Hellsten,
  Jaakko Lehtinen, and Timo Aila.
\newblock Alias-free generative adversarial networks.
\newblock In {\em Proc. NeurIPS}, 2021.

\bibitem{karras2018styleGANv1}
Tero Karras, Samuli Laine, and Timo Aila.
\newblock A style-based generator architecture for generative adversarial
  networks.
\newblock In {\em Proceedings of the IEEE/CVF conference on computer vision and
  pattern recognition}, pages 4401--4410, 2019.

\bibitem{karras2020styleganv2}
Tero Karras, Samuli Laine, Miika Aittala, Janne Hellsten, Jaakko Lehtinen, and
  Timo Aila.
\newblock Analyzing and improving the image quality of stylegan.
\newblock In {\em Proceedings of the IEEE/CVF Conference on Computer Vision and
  Pattern Recognition}, pages 8110--8119, 2020.

\bibitem{kingma2014adam}
Diederik~P Kingma and Jimmy Ba.
\newblock Adam: A method for stochastic optimization.
\newblock In {\em ICLR}, 2015.

\bibitem{kirkpatrick2017ewc_pnas}
James Kirkpatrick, Razvan Pascanu, Neil Rabinowitz, Joel Veness, Guillaume
  Desjardins, Andrei~A Rusu, Kieran Milan, John Quan, Tiago Ramalho, Agnieszka
  Grabska-Barwinska, et~al.
\newblock Overcoming catastrophic forgetting in neural networks.
\newblock {\em Proceedings of the national academy of sciences},
  114(13):3521--3526, 2017.

\bibitem{percy:finetuning:iclr22}
Ananya Kumar, Aditi Raghunathan, Robbie Jones, Tengyu Ma, and Percy Liang.
\newblock Fine-tuning can distort pretrained features and underperform
  out-of-distribution.
\newblock In {\em ICLR}, 2022.

\bibitem{lacoste2019quantifying_co2}
Alexandre Lacoste, Alexandra Luccioni, Victor Schmidt, and Thomas Dandres.
\newblock Quantifying the carbon emissions of machine learning.
\newblock {\em arXiv preprint arXiv:1910.09700}, 2019.

\bibitem{lihonglin2023task}
Honglin Li, Chenglu Zhu, Yunlong Zhang, Yuxuan Sun, Zhongyi Shui, Wenwei Kuang,
  Sunyi Zheng, and Lin Yang.
\newblock Task-specific fine-tuning via variational information bottleneck for
  weakly-supervised pathology whole slide image classification.
\newblock {\em arXiv preprint arXiv:2303.08446}, 2023.

\bibitem{li2022source}
Shuaifeng Li, Mao Ye, Xiatian Zhu, Lihua Zhou, and Lin Xiong.
\newblock Source-free object detection by learning to overlook domain style.
\newblock In {\em Proceedings of the IEEE/CVF Conference on Computer Vision and
  Pattern Recognition}, pages 8014--8023, 2022.

\bibitem{li_new2022dynamic}
Tianjiao Li, Lin~Geng Foo, Qiuhong Ke, Hossein Rahmani, Anran Wang, Jinghua
  Wang, and Jun Liu.
\newblock Dynamic spatio-temporal specialization learning for fine-grained
  action recognition.
\newblock In {\em Computer Vision--ECCV 2022: 17th European Conference, Tel
  Aviv, Israel, October 23--27, 2022, Proceedings, Part IV}, pages 386--403.
  Springer, 2022.

\bibitem{li_new2021else}
Tianjiao Li, Qiuhong Ke, Hossein Rahmani, Rui~En Ho, Henghui Ding, and Jun Liu.
\newblock Else-net: Elastic semantic network for continual action recognition
  from skeleton data.
\newblock In {\em Proceedings of the IEEE/CVF International Conference on
  Computer Vision}, pages 13434--13443, 2021.

\bibitem{li_new2020hard}
Tianjiao Li, Jun Liu, Wei Zhang, and Lingyu Duan.
\newblock Hard-net: Hardness-aware discrimination network for 3d early activity
  prediction.
\newblock In {\em Computer Vision--ECCV 2020: 16th European Conference,
  Glasgow, UK, August 23--28, 2020, Proceedings, Part XI 16}, pages 420--436.
  Springer, 2020.

\bibitem{li_new2021uav}
Tianjiao Li, Jun Liu, Wei Zhang, Yun Ni, Wenqian Wang, and Zhiheng Li.
\newblock Uav-human: A large benchmark for human behavior understanding with
  unmanned aerial vehicles.
\newblock In {\em Proceedings of the IEEE/CVF conference on computer vision and
  pattern recognition}, pages 16266--16275, 2021.

\bibitem{li_new2021pot}
Tianjiao Li, Wei Zhang, Ran Song, Zhiheng Li, Jun Liu, Xiaolei Li, and Shijian
  Lu.
\newblock Pot-gan: pose transform gan for person image synthesis.
\newblock {\em IEEE Transactions on Image Processing}, 30:7677--7688, 2021.

\bibitem{li:inductive:icml18}
Xuhong LI, Yves GRANDVALET, and Franck DAVOINE.
\newblock Explicit inductive bias for transfer learning with convolutional
  networks.
\newblock In {\em ICML}, 2018.

\bibitem{li2020fig_EWC}
Yijun Li, Richard Zhang, Jingwan~(Cynthia) Lu, and Eli Shechtman.
\newblock Few-shot image generation with elastic weight consolidation.
\newblock In H. Larochelle, M. Ranzato, R. Hadsell, M.~F. Balcan, and H. Lin,
  editors, {\em Advances in Neural Information Processing Systems}, volume~33,
  pages 15885--15896. Curran Associates, Inc., 2020.

\bibitem{li2020train}
Zhuohan Li, Eric Wallace, Sheng Shen, Kevin Lin, Kurt Keutzer, Dan Klein, and
  Joey Gonzalez.
\newblock Train big, then compress: Rethinking model size for efficient
  training and inference of transformers.
\newblock In {\em International Conference on Machine Learning}, pages
  5958--5968. PMLR, 2020.

\bibitem{liang2021finding}
Jianze Liang, Chengqi Zhao, Mingxuan Wang, Xipeng Qiu, and Lei Li.
\newblock Finding sparse structures for domain specific neural machine
  translation.
\newblock In {\em Proceedings of the AAAI Conference on Artificial
  Intelligence}, volume~35, pages 13333--13342, 2021.

\bibitem{lin2017runtime}
Ji Lin, Yongming Rao, Jiwen Lu, and Jie Zhou.
\newblock Runtime neural pruning.
\newblock {\em Advances in neural information processing systems}, 30, 2017.

\bibitem{lin2021anycostGANs}
Ji Lin, Richard Zhang, Frieder Ganz, Song Han, and Jun-Yan Zhu.
\newblock Anycost gans for interactive image synthesis and editing.
\newblock In {\em Proceedings of the IEEE/CVF Conference on Computer Vision and
  Pattern Recognition}, pages 14986--14996, 2021.

\bibitem{gong2022system}
Foo Lin~Geng, Jia Gong, Zhipeng Fan, and Jun Liu.
\newblock System-status-aware adaptive network for online streaming video
  understanding.
\newblock In {\em Proceedings of the IEEE/CVF Conference on Computer Vision and
  Pattern Recognition}, 2023.

\bibitem{mika1999fisher}
Sebastian Mika, Gunnar Ratsch, Jason Weston, Bernhard Scholkopf, and
  Klaus-Robert Mullers.
\newblock Fisher discriminant analysis with kernels.
\newblock In {\em Neural networks for signal processing IX: Proceedings of the
  1999 IEEE signal processing society workshop (cat. no. 98th8468)}, pages
  41--48. Ieee, 1999.

\bibitem{mo2020freezeD}
Sangwoo Mo, Minsu Cho, and Jinwoo Shin.
\newblock Freeze the discriminator: a simple baseline for fine-tuning gans.
\newblock In {\em CVPR AI for Content Creation Workshop}, 2020.

\bibitem{noguchi2019BSA}
Atsuhiro Noguchi and Tatsuya Harada.
\newblock Image generation from small datasets via batch statistics adaptation.
\newblock In {\em Proceedings of the IEEE/CVF International Conference on
  Computer Vision}, pages 2750--2758, 2019.

\bibitem{ojha2021fig_cdc}
Utkarsh Ojha, Yijun Li, Jingwan Lu, Alexei~A Efros, Yong~Jae Lee, Eli
  Shechtman, and Richard Zhang.
\newblock Few-shot image generation via cross-domain correspondence.
\newblock In {\em Proceedings of the IEEE/CVF Conference on Computer Vision and
  Pattern Recognition}, pages 10743--10752, 2021.

\bibitem{pan2009yang-qiang-transfer}
Sinno~Jialin Pan and Qiang Yang.
\newblock A survey on transfer learning.
\newblock {\em IEEE Transactions on knowledge and data engineering},
  22(10):1345--1359, 2009.

\bibitem{paszke2019pytorch}
Adam Paszke, Sam Gross, Francisco Massa, Adam Lerer, James Bradbury, Gregory
  Chanan, Trevor Killeen, Zeming Lin, Natalia Gimelshein, Luca Antiga, et~al.
\newblock Pytorch: An imperative style, high-performance deep learning library.
\newblock {\em Advances in neural information processing systems},
  32:8026--8037, 2019.

\bibitem{rezende2015variational}
Danilo Rezende and Shakir Mohamed.
\newblock Variational inference with normalizing flows.
\newblock In {\em International conference on machine learning}, pages
  1530--1538. PMLR, 2015.

\bibitem{rombach2021highresolution}
Robin Rombach, Andreas Blattmann, Dominik Lorenz, Patrick Esser, and Björn
  Ommer.
\newblock High-resolution image synthesis with latent diffusion models, 2021.

\bibitem{Sermanet:Overfeat:iclr2014}
P. Sermanet, D. Eigen, X. Zhang, M. Mathieu, R. Fergus, and Y. LeCun.
\newblock Overfeat: Integrated recognition, localization and detection using
  convolutional networks.
\newblock In {\em ICLR}, 2014.

\bibitem{shen2022prune}
Maying Shen, Pavlo Molchanov, Hongxu Yin, and Jose~M Alvarez.
\newblock When to prune? a policy towards early structural pruning.
\newblock In {\em Proceedings of the IEEE/CVF Conference on Computer Vision and
  Pattern Recognition}, pages 12247--12256, 2022.

\bibitem{simonyan2013class_salience}
Karen Simonyan, Andrea Vedaldi, and Andrew Zisserman.
\newblock Deep inside convolutional networks: Visualising image classification
  models and saliency maps.
\newblock {\em arXiv preprint arXiv:1312.6034}, 2013.

\bibitem{simonyan2013deep}
Karen Simonyan, Andrea Vedaldi, and Andrew Zisserman.
\newblock Deep inside convolutional networks: Visualising image classification
  models and saliency maps.
\newblock {\em arXiv preprint arXiv:1312.6034}, 2013.

\bibitem{sui2021chip}
Yang Sui, Miao Yin, Yi Xie, Huy Phan, Saman Aliari~Zonouz, and Bo Yuan.
\newblock Chip: Channel independence-based pruning for compact neural networks.
\newblock {\em Advances in Neural Information Processing Systems},
  34:24604--24616, 2021.

\bibitem{sun2020explanation}
Jiamei Sun, Sebastian Lapuschkin, Wojciech Samek, Yunqing Zhao, Ngai-Man
  Cheung, and Alexander Binder.
\newblock Explanation-guided training for cross-domain few-shot classification.
\newblock {\em arXiv preprint arXiv:2007.08790}, 2020.

\bibitem{tran2021data_aug_gan}
Ngoc-Trung Tran, Viet-Hung Tran, Ngoc-Bao Nguyen, Trung-Kien Nguyen, and
  Ngai-Man Cheung.
\newblock On data augmentation for gan training.
\newblock {\em IEEE Transactions on Image Processing}, 30:1882--1897, 2021.

\bibitem{tseng2021regularizingGAN}
Hung-Yu Tseng, Lu Jiang, Ce Liu, Ming-Hsuan Yang, and Weilong Yang.
\newblock Regularizing generative adversarial networks under limited data.
\newblock In {\em Proceedings of the IEEE/CVF Conference on Computer Vision and
  Pattern Recognition}, pages 7921--7931, 2021.

\bibitem{van2017vq_vae}
Aaron Van Den~Oord, Oriol Vinyals, et~al.
\newblock Neural discrete representation learning.
\newblock {\em Advances in neural information processing systems}, 30, 2017.

\bibitem{wang2020apq}
Tianzhe Wang, Kuan Wang, Han Cai, Ji Lin, Zhijian Liu, Hanrui Wang, Yujun Lin,
  and Song Han.
\newblock Apq: Joint search for network architecture, pruning and quantization
  policy.
\newblock In {\em Proceedings of the IEEE/CVF Conference on Computer Vision and
  Pattern Recognition}, pages 2078--2087, 2020.

\bibitem{wang2022high}
Tengfei Wang, Yong Zhang, Yanbo Fan, Jue Wang, and Qifeng Chen.
\newblock High-fidelity gan inversion for image attribute editing.
\newblock In {\em Proceedings of the IEEE/CVF Conference on Computer Vision and
  Pattern Recognition}, pages 11379--11388, 2022.

\bibitem{wang2008cuhk_sketches}
Xiaogang Wang and Xiaoou Tang.
\newblock Face photo-sketch synthesis and recognition.
\newblock {\em IEEE transactions on pattern analysis and machine intelligence},
  31(11):1955--1967, 2008.

\bibitem{wang2020minegan}
Yaxing Wang, Abel Gonzalez-Garcia, David Berga, Luis Herranz, Fahad~Shahbaz
  Khan, and Joost van~de Weijer.
\newblock Minegan: effective knowledge transfer from gans to target domains
  with few images.
\newblock In {\em Proceedings of the IEEE/CVF Conference on Computer Vision and
  Pattern Recognition}, pages 9332--9341, 2020.

\bibitem{wang2018transferringGAN}
Yaxing Wang, Chenshen Wu, Luis Herranz, Joost van~de Weijer, Abel
  Gonzalez-Garcia, and Bogdan Raducanu.
\newblock Transferring gans: generating images from limited data.
\newblock In {\em Proceedings of the European Conference on Computer Vision
  (ECCV)}, pages 218--234, 2018.

\bibitem{prune_redundancy}
Zi Wang, Chengcheng Li, and Xiangyang Wang.
\newblock Convolutional neural network pruning with structural redundancy
  reduction.
\newblock In {\em Proceedings of the IEEE/CVF Conference on Computer Vision and
  Pattern Recognition (CVPR)}, pages 14913--14922, June 2021.

\bibitem{webster2019detecting_overfit_fid}
Ryan Webster, Julien Rabin, Loic Simon, and Fr{\'e}d{\'e}ric Jurie.
\newblock Detecting overfitting of deep generative networks via latent
  recovery.
\newblock In {\em Proceedings of the IEEE/CVF Conference on Computer Vision and
  Pattern Recognition}, pages 11273--11282, 2019.

\bibitem{xiao2022few}
Jiayu Xiao, Liang Li, Chaofei Wang, Zheng-Jun Zha, and Qingming Huang.
\newblock Few shot generative model adaption via relaxed spatial structural
  alignment.
\newblock {\em arXiv preprint arXiv:2203.04121}, 2022.

\bibitem{yanbo_2022_CVPR_editing}
Yanbo Xu, Yueqin Yin, Liming Jiang, Qianyi Wu, Chengyao Zheng, Chen~Change Loy,
  Bo Dai, and Wayne Wu.
\newblock Transeditor: Transformer-based dual-space gan for highly controllable
  facial editing.
\newblock In {\em Proceedings of the IEEE/CVF Conference on Computer Vision and
  Pattern Recognition (CVPR)}, pages 7683--7692, June 2022.

\bibitem{yang2021InsGen}
Ceyuan Yang, Yujun Shen, Yinghao Xu, and Bolei Zhou.
\newblock Data-efficient instance generation from instance discrimination.
\newblock {\em Advances in Neural Information Processing Systems}, 34, 2021.

\bibitem{yang2021one-shot-adaptation}
Ceyuan Yang, Yujun Shen, Zhiyi Zhang, Yinghao Xu, Jiapeng Zhu, Zhirong Wu, and
  Bolei Zhou.
\newblock One-shot generative domain adaptation.
\newblock {\em arXiv preprint arXiv:2111.09876}, 2021.

\bibitem{yaniv2019face_Modigliani}
Jordan Yaniv, Yael Newman, and Ariel Shamir.
\newblock The face of art: landmark detection and geometric style in portraits.
\newblock {\em ACM Transactions on graphics (TOG)}, 38(4):1--15, 2019.

\bibitem{Yosinski:transfer:nips14}
J. Yosinski, J. Clune, Y. Bengio, and H. Lipson.
\newblock How transferable are features in deep neural networks?
\newblock In {\em NIPS}, 2014.

\bibitem{yu15lsun}
Fisher Yu, Yinda Zhang, Shuran Song, Ari Seff, and Jianxiong Xiao.
\newblock Lsun: Construction of a large-scale image dataset using deep learning
  with humans in the loop.
\newblock {\em arXiv preprint arXiv:1506.03365}, 2015.

\bibitem{Fergus:visualize:2013}
Matthew~D. Zeiler and Rob Fergus.
\newblock Visualizing and understanding convolutional networks.
\newblock {\em CoRR}, abs/1311.2901, 2013.

\bibitem{zhang2018lpips}
Richard Zhang, Phillip Isola, Alexei~A Efros, Eli Shechtman, and Oliver Wang.
\newblock The unreasonable effectiveness of deep features as a perceptual
  metric.
\newblock In {\em CVPR}, 2018.

\bibitem{zhao2020leveraging_icml_adafm}
Miaoyun Zhao, Yulai Cong, and Lawrence Carin.
\newblock On leveraging pretrained gans for generation with limited data.
\newblock In {\em International Conference on Machine Learning}, pages
  11340--11351. PMLR, 2020.

\bibitem{zhao2022fsig-ip}
Yunqing Zhao, Keshigeyan Chandrasegaran, Milad Abdollahzadeh, and Ngai man
  Cheung.
\newblock Few-shot image generation via adaptation-aware kernel modulation.
\newblock In {\em Thirty-Sixth Conference on Neural Information Processing
  Systems}, 2022.

\bibitem{Zhao_2023_tip_fsc}
Yunqing Zhao and Ngai-Man Cheung.
\newblock Fs-ban: Born-again networks for domain generalization few-shot
  classification.
\newblock {\em IEEE Transactions on Image Processing}, 2023.

\bibitem{zhao2022dcl}
Yunqing Zhao, Henghui Ding, Houjing Huang, and Ngai-Man Cheung.
\newblock A closer look at few-shot image generation.
\newblock In {\em CVPR}, 2022.

\bibitem{Zhao_2023_arxiv_watermark_dm}
Yunqing Zhao, Tianyu Pang, Chao Du, Xiao Yang, Ngai-Man Cheung, and Min Lin.
\newblock A recipe for watermarking diffusion models.
\newblock {\em arXiv preprint arXiv: 2303.10137}, 2023.

\bibitem{zhou2017places}
Bolei Zhou, Agata Lapedriza, Aditya Khosla, Aude Oliva, and Antonio Torralba.
\newblock Places: A 10 million image database for scene recognition.
\newblock {\em IEEE transactions on pattern analysis and machine intelligence},
  40(6):1452--1464, 2017.

\bibitem{zhu2020domain}
Jiapeng Zhu, Yujun Shen, Deli Zhao, and Bolei Zhou.
\newblock In-domain gan inversion for real image editing.
\newblock In {\em European conference on computer vision}, pages 592--608.
  Springer, 2020.

\bibitem{zhuang:transfer:IEEE2021}
Fuzhen Zhuang, Zhiyuan Qi, Keyu Duan, Dongbo Xi, Yongchun Zhu, Hengshu Zhu, Hui
  Xiong, and Qing He.
\newblock A comprehensive survey on transfer learning.
\newblock {\em Proceedings of the IEEE}, 109(1):43--76, 2021.

\end{thebibliography}
}

\clearpage

{
\onecolumn
\appendix

\renewcommand{\thetable}{S\arabic{table}}
\renewcommand{\thefigure}{S\arabic{figure}}

\section*{Overview}
 This supplementary material provides the  additional experiments  and  results  to further support our main findings and proposed method for few-shot image generation (FSIG).
 These were not included in the main paper due to the space limitations.
 % \textbf{Reproducibility.} We provide the {\color{RubineRed}{Code}} to help reproduce the experiments in our work. Please refer to the following anonymous links:
 % \begin{itemize}
 %    \setlength{\itemsep}{8pt}
 %    % \setlength{\topsep}{5pt}
 %    \setlength{\parsep}{1pt}
 %    \setlength{\parskip}{1pt}
 %     \item \textbf{Code:} \href{https://drive.google.com/file/d/17EWBzMo8LiEfGhvzjCeE7dukW5xEeFSR/view?usp=share_link}{\textbf{[Code]}}
 %     \item \textbf{Datasets:}
 %     \href{https://drive.google.com/drive/folders/19qQDo4xVm5G9VYyx_cRlSSW7jFTDh7VC?usp=share_link}{\textbf{[Datasets]}}
 %     \item \textbf{Pretrained Models:}
 %     \href{https://drive.google.com/drive/folders/1gKztAnglu11As27WB1Lhy5cnnQUkqaRz?usp=share_link}{\textbf{[Pretrained Models]}}
 % \end{itemize}
 The supplementary material is organized as follows:
{
    \hypersetup{linkcolor=black}
    \tableofcontents
    % \listoffigures
    % \listoftables
}

\newpage

\section{Pesudo-code of our proposed method}
\label{sec-algo}
%------------------------------------------
Here, we describe the proposed method in algorithmic format. We use the generator $G$ for example, but the same operations are also applied to the discriminator $D$ during adaptation. 
As mentioned in Sec. {\color{red} 5.1} in the main paper, we use Fisher Information (FI) as the measurement for importance estimation in main experiments, and we use the same notation in the algorithm. 
% Note that the weights of linear layers ($\mathbf{W}_{linear}\in \mathbb{R}^{c^{out}\times c^{in}}$) are also \textit{preserved}, \textit{fine-tuned} or \textit{pruned}, similar to the conv layers. 
% In this section, we use conv filters for illustration.
We summarize the proposed method in Algorithm \ref{alg:method_algo}.

\begin{algorithm}[t]
\caption{Proposed Knowledge Truncation for Few-shot Image Generation}
\label{alg:method_algo}
\begin{algorithmic}[1]
\STATE  \textbf{Input:} target samples   $\mathbf{X}$, number of total iterations $\mathbf{N}$,
number of warmup iterations $\mathbf{N}_{w}$, interval for filter importance estimation $\mathbf{T}$, Conv filters of the model $\mathbf{W}$, FI of the filter $\mathcal{F}(\mathbf{W})$, quantile of filter importance $q(\mathbf{W})$, quantile threshold for filter that is selected for pruning $t_{l}$, threshold for filter that is selected for preserving $t_{h}$, total pruning rate $p\%$.

\STATE  \textbf{Initialize with pretrained GAN}: $G_{t} \gets G_{s}$ and $D_{t} \gets D_{s}$.
\FOR{$iteration=1$; $iteration \leq \mathbf{N}_{w}$; $iteration++$}
	\STATE Update the discriminator $D_{t}$ given $\mathbf{X}$, via Eqn. {\color{red} 1} {\color{gray} \texttt{ \# warmup for $D_{t}$}}
\ENDFOR
\FOR{$iteration=\mathbf{N}_{w}+1$; $iteration \leq \mathbf{N}$; $iteration++$}
    \STATE \textbf{If} $iteration\%\mathbf{T}=0:$ \\ \quad Estimate $\mathcal{F}(\mathbf{W})$ via Eqn. {\color{red} 2};\\
    \quad Preserve $\mathbf{W}$ if $t_{h} \leq q(\mathbf{W})$;\\
    \quad Fine-tune $\mathbf{W}$ if $t_{l} \leq q(\mathbf{W}) \leq t_{h}$;\\
    \quad Prune $\mathbf{W}$ if $q(\mathbf{W}) \leq t_{l}$;\\
    \quad Update the operation of $\mathbf{W}$ in Memory Bank. {\color{gray} \texttt{\# size of the Memory Bank $\sim$ 5,000}}
    \STATE \textbf{ELSE}\\
    \quad Extract the operation of $\mathbf{W}$ in Memory Bank;\\
    \quad Update model via Eqn. {\color{red} 1}.
\ENDFOR
\STATE \textbf{Output:} The adapted GAN
\end{algorithmic} 
\end{algorithm}

%-------------------------------------------------------

\section{Pseudo-code for implementing Intra-LPIPS}
\label{sec-lpips}

Here we describe the implementation details of Intra-LPIPS, the diversity evaluation method proposed in \cite{ojha2021fig_cdc}. 
We follow prior works \cite{ojha2021fig_cdc, zhao2022dcl, zhao2022fsig-ip} to apply Intra-LPIPS as an additional metric to evaluate the diversity of generated images by different adapted GAN models. 
To make it easy to understand, we summarize the pseudo-code of Intra-LPIPS in Algorithm \ref{intra-lpips-algo}.
A small value of Intra-LPIPS indicates that the generated images are more close to few-shot target samples.

%%%%%%%%%%%%%%%%%%%%%%%%%%%%%%%%%%%
%%%%%%%%%%%%%%%%%%%%%%%%%%%%%%%%%%%
%%%%%%%%%%%%%%%%%%%%%%%%%%%%%%%%%%%
\begin{algorithm}[t]
\caption{Pseudo-code of Intra-LPIPS} 
\definecolor{codegreen}{rgb}{0,0.6,0}
\definecolor{codegray}{rgb}{0.5,0.5,0.5}
\definecolor{codepurple}{rgb}{0.58,0,0.82}
\definecolor{backcolour}{rgb}{0.99,0.99,0.99}

\lstdefinestyle{mystyle}{
    backgroundcolor=\color{backcolour},   
    commentstyle=\color{codegray},
    keywordstyle=\color{magenta},
    numberstyle=\tiny\color{codegray},
    stringstyle=\color{codepurple},
    basicstyle=\ttfamily\footnotesize,
    breakatwhitespace=false,         
    breaklines=true,                 
    captionpos=b,                    
    keepspaces=true,                 
    numbers=left,                    
    numbersep=5pt,                  
    showspaces=false,                
    showstringspaces=false,
    showtabs=false,                  
    tabsize=2
}
\lstset{style=mystyle}
\label{intra-lpips-algo}
\begin{lstlisting}[language=Python]
# Input: 1. Generated images X=[x1, ..., xn]; 
#        2. Suppose we have 2-shot target samples with cluster center: c0, c1; 
#        3. Cluster_0, Cluster_1 = [], []
# Output: Avg Intra-LPIPS over 2 clusters
# ------------------------------------------- #
# Step 0. Define the LPIPS distance function (Zhang et al.)
import lpips

lpips_fn = lpips.LPIPS(net='vgg')  # default setup

# Step 1. Assign generated images to the cluster with smallest LPIPS distance to the cluster center
for X[i] in X:
    dist0 = lpips_fn(X[i], c0)
    dist1 = lpips_fn(X[i], c1)
    if dist0 < dist1:
        Cluster_0.append(X[i])
    else:
        Cluster_1.append(X[i])
# ------------------------------------------- #

# Step 2. Compute Intra-LPIPS
lpips_dist = []
While not done:  # randomly sample image pairs
    for img_i, img_j in Cluster_0:
        lpips_dist.append(lpips_fn(img_i, img_j))
    for img_i, img_j in Cluster_1:
        lpips_dist.append(lpips_fn(img_i, img_j))
return lpips_dist.mean()
# ------------------------------------------- #

\end{lstlisting}
\end{algorithm}
%%%%%%%%%%%%%%%%%%%%%%%%%%%%%
%%%%%%%%%%%%%%%%%%%%%%%%%%%%%
%%%%%%%%%%%%%%%%%%%%%%%%%%%%%

\section{A brief description of Importance Probing (IP)}
\label{sec-intro-ip}

In the main paper, we apply IP, a modulation based method that estimates the filter importance for target adaptation proposed in AdAM \cite{zhao2022fsig-ip}, to determine the filters importance for target adaptation. We discover that the filters with least importance for adaptation are highly correlated with the knowledge that is incompatible to the target domain.

In this section, we briefly discuss the implementation of IP.
We refer the readers to~\cite{zhao2022fsig-ip} for more details.
In IP, for each filter in the generator $\mathbf{W}$, they generate a new filter $\mathbf{M}$ with the same dimension as $\mathbf{W}$ to modulate $\mathbf{W}$. 
Then, the modulated filter $\mathbf{W}^{'}$ can be written as follows:
\begin{equation}
    \mathbf{W}^{'} = \mathbf{W} \odot (\mathbf{J} + \mathbf{M}),
\end{equation}
where $\mathbf{J}$ is the all-one matrix, $\odot$ indicates the Hadamard product.
Then, they adapt the modulation matrix $\mathbf{W}^{'}$ to target domain for 500 iterations (called ``probing'') and only update $\mathbf{M}$ during the probing stage. After that, they use the importance (estimated) of modulation matrix $\mathbf{M}$ as approximation of importance for the original filters $\mathbf{W}$.  

In contrast to IP, we directly estimate the filter importance for adaptation by leveraging on-the-fly gradient information during training (therefore no probing stage). The results in Table {\color{red} 2} in the main paper shows we can achieve comparable performance even without our proposed knowledge truncation method.

\section{Additional experiment details}
\label{sec-implementation-detail}
Here we provide additional details to help reproduce our results.

\textbf{Code Implementation.}
 For GAN dissection experiments and analysis, we follow their implementation and use the official \href{https://github.com/CSAILVision/gandissect}{repo}. 
 In main experiments, following prior SOTA methods \cite{ojha2021fig_cdc, zhao2022dcl, zhao2022fsig-ip}, we use StyleGAN-V2 as GAN architecture for a fair comparison.
 The implementation is from this \href{https://github.com/rosinality/stylegan2-pytorch}{repo}. 
 We additionally use the ProgressiveGAN \cite{karras2017progan} to validate our proposed analysis and method, using the code base from Karras \etal \cite{karras2017progan}. 
 For performance evaluation, we use the tool (\eg, to compute FID) used in \cite{cong2020gan_memory}.
 During adaptation, we strictly follow the default hyperparameters as prior works \cite{ojha2021fig_cdc, zhao2022dcl}.
 Before the main adaptation, there is a warmup stage where we only update $D_{t}$ for a few iterations (\eg, 250 in our experiments) to let the $D_{t}$ aware of the target domain information to have a better estimation of the filter importance for the target adaptation. 
 We evaluate the filter importance for adaptation every 50 iterations. 
 Nevertheless, we found our proposed method is not very sensitive to the choice of these hyperparameters.
 We use batch size 4 and an initial learning rate 0.002 (with Cosine Scheduler in PyTorch \cite{paszke2019pytorch}) in all experiments, similar to prior works \cite{ojha2021fig_cdc, zhao2022dcl, zhao2022fsig-ip}. We use images with resolution 256 x 256 for adaptation.

\textbf{Datasets.} 
 In main paper, we follow prior works \cite{li2020fig_EWC, ojha2021fig_cdc, zhao2022dcl, zhao2022fsig-ip} in the choice of source and target datasets for adaptation. 
 To make a fair and comprehensive comparison, we use FFHQ \cite{karras2018styleGANv1} as source domain. 
 In this supplementary material, we additionally use LSUN Church \cite{yu15lsun} as source domain for visualization. We also include more target datasets for adaptation, \eg, face paintings \cite{yaniv2019face_Modigliani, crockett1992face_otto_dix, wang2008cuhk_sketches}, Haunted Houses or Palace \cite{deng2009imagenet}.
 % to be added.
 We perform 10-shot adaptation in most setups. Nevertheless, we also evaluate our method given different number of target training samples, see results in Table \ref{table:supp-fid-shot}.

% \textbf{Conda environment information.}
% We note that crash may happen between python packages if simply follow the guidelines in \cite{bau2019gan}, due to some packages and CUDA versions are outdated.
% To help reproduce the main analysis, \eg, GAN dissection \cite{bau2019gan} in Figure {\color{red} 2}, and experiment results in Table {\color{red} 1} and Figure {\color{red} 4}, 
% To help reproduce the results in our paper, \eg, Table {\color{red} 1} and Figure {\color{red} 4},
% we provide the conda environment configurations, and it can be achieved in this \href{https://drive.google.com/file/d/1N2jhAo8ROXNtgB0rb_pVDa0_JoxLmjUs/view?usp=share_link}{anonymous link}. See details in the code submission.

\textbf{Implementation details of pruning EWC and AdAM.}
In literature, EWC \cite{li2020fig_EWC} and AdAM \cite{zhao2022fsig-ip} proposed different criteria to preserve source knowledge by identifying parameter importance. 
EWC (Li \etal) applied Fisher Information (FI) to measure parameter importance to the source domain. They then penalize the change of parameters during adaptation, weighted by the importance of the parameters: if a parameter is 
deemed to be important, it will be penalized more given the same change. 
AdAM proposed a modulation based method to evaluate the parameter importance for adaptation, see details in Sec. \ref{sec-intro-ip}.
%prune
Therefore, in Table {\color{red} 1} in the main paper, we propose to prune least important filters using the importance measurement proposed in their original works, and empirically we observed the improved FID score of the generated images, which implies the effectiveness and generalizability of our proposed method.

\begin{figure}[t]
    \centering
    \includegraphics[width=0.88\textwidth]{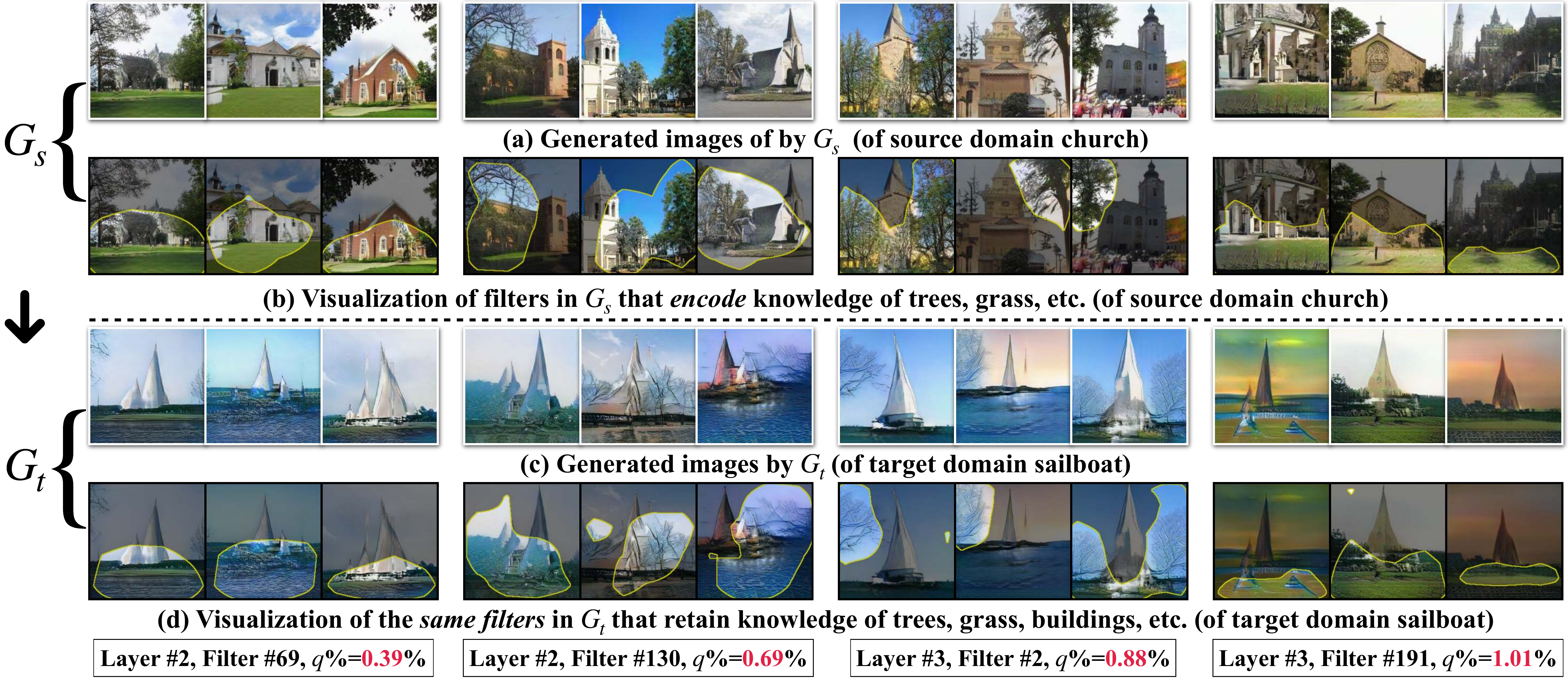}
    \includegraphics[width=0.88\textwidth]{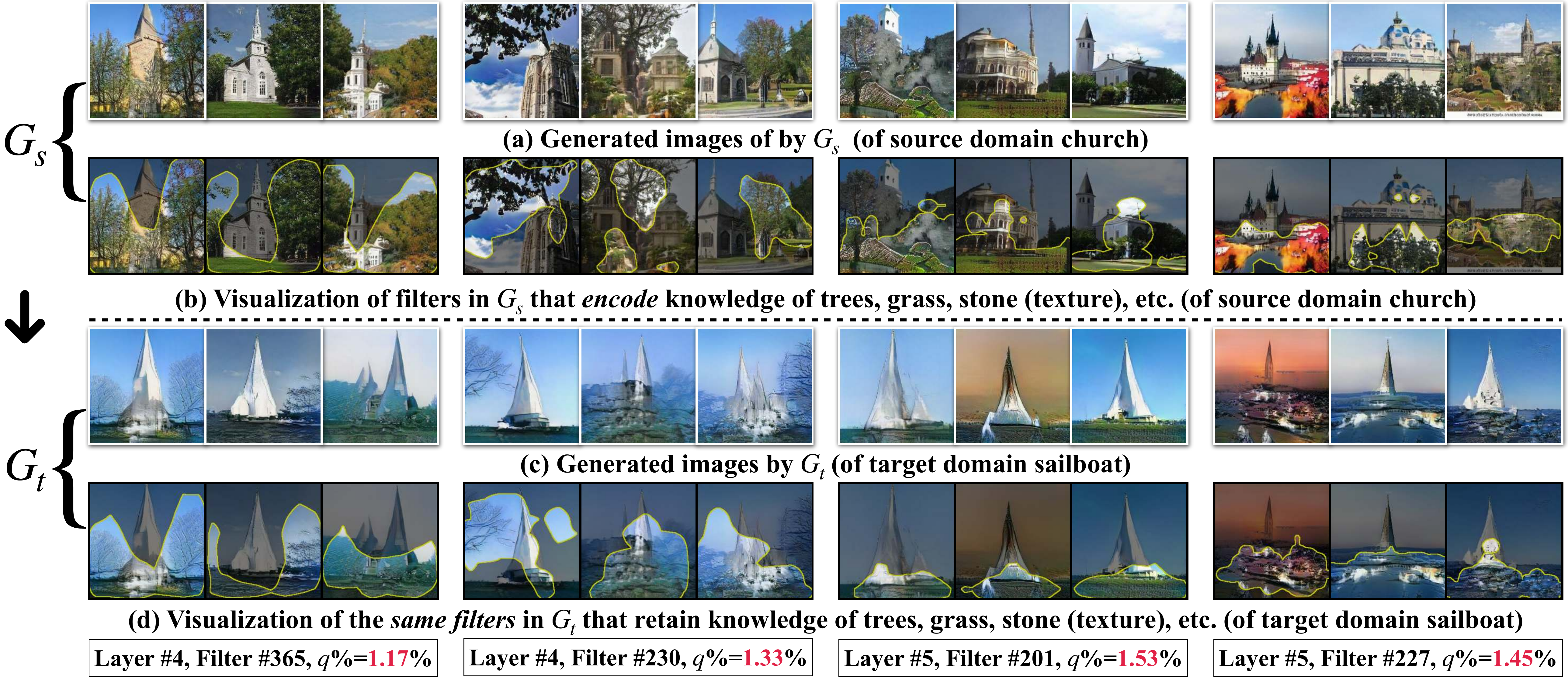}
    \includegraphics[width=0.88\textwidth]{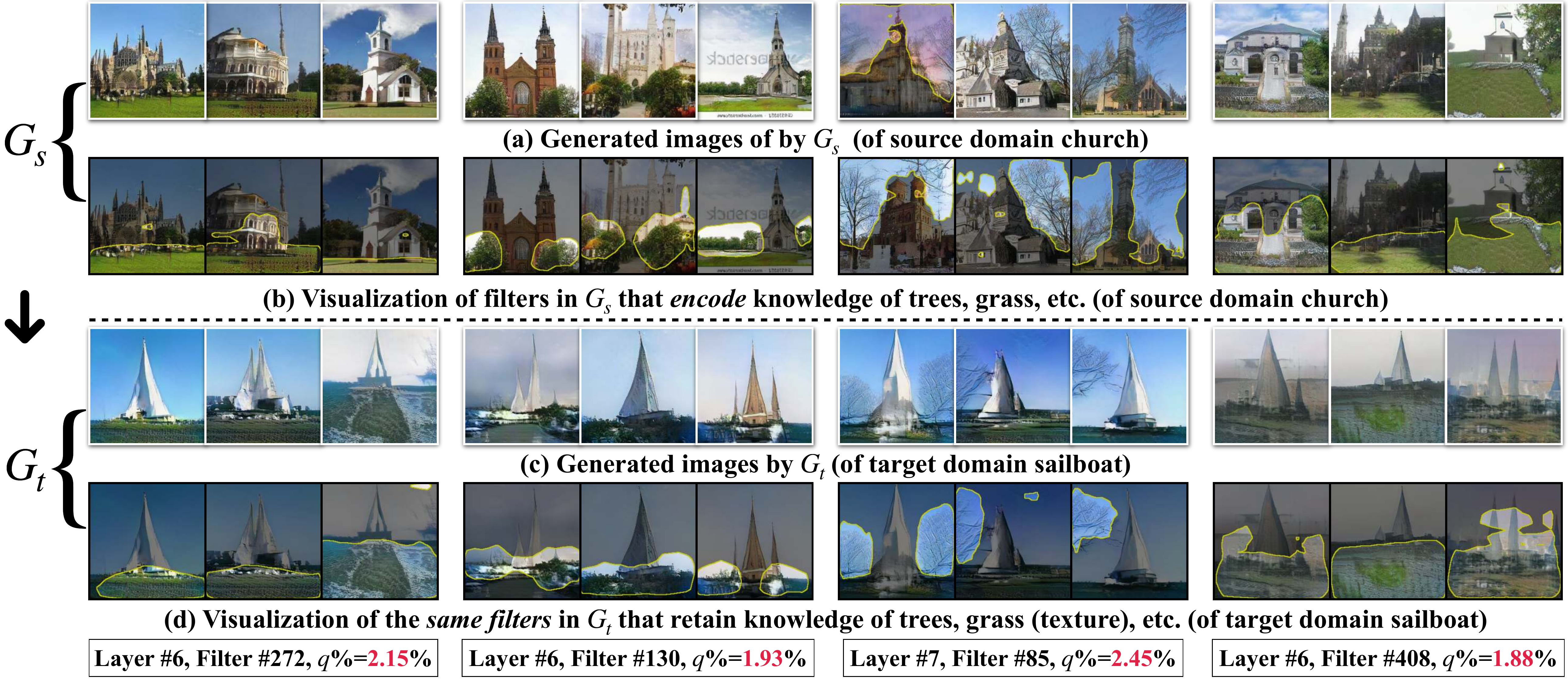}
    \caption{Additional GAN dissection results using AdAM.
    Note that in AdAM fine-tuning is applied during adaptation from $G_s$ to $G_t$ to update these low importance filters. However, we observe that incompatible knowledge (tree, building, grass) remains in the same filters in $G_t$ after fine-tuning.
    }
    \label{fig:diss_supp_adam}
\end{figure}

\begin{figure}[t]
    \centering
    \includegraphics[width=0.88\textwidth]{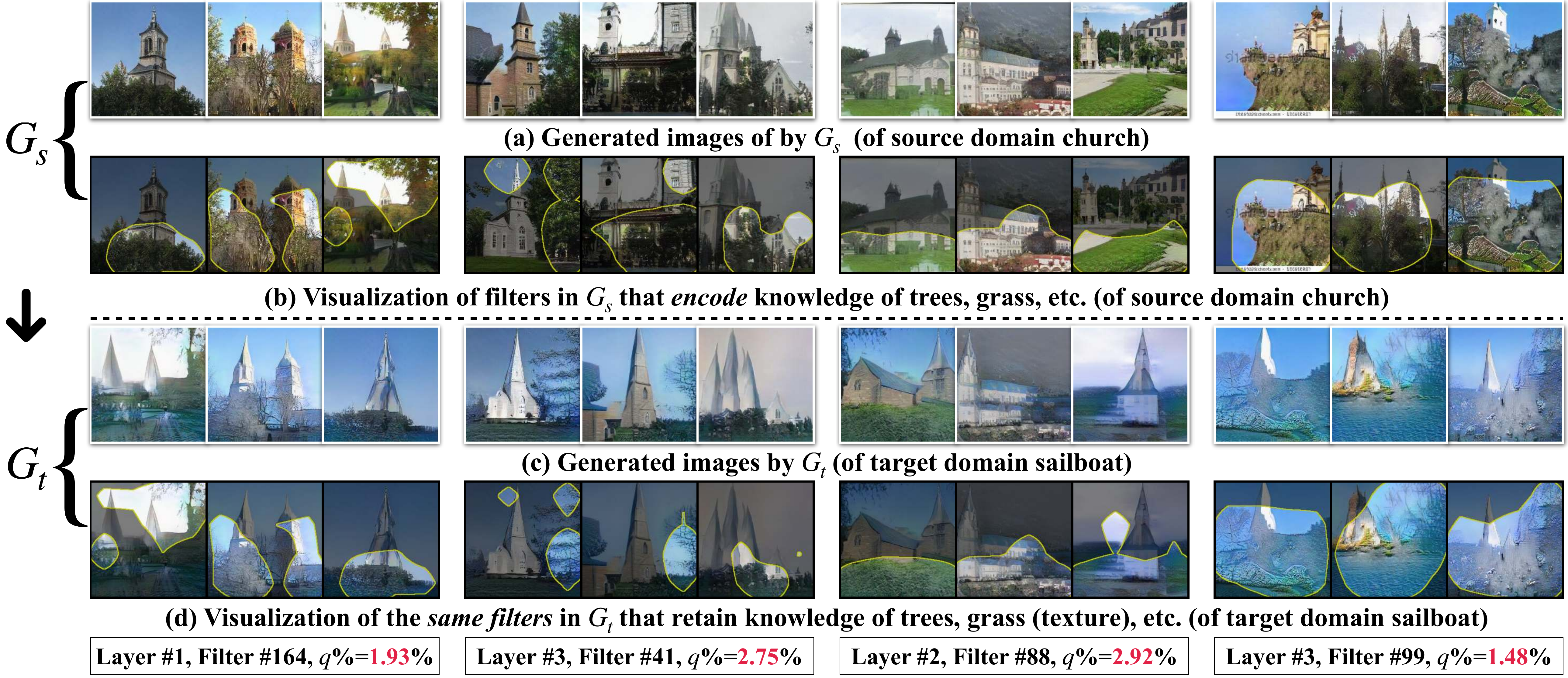}
    \includegraphics[width=0.88\textwidth]{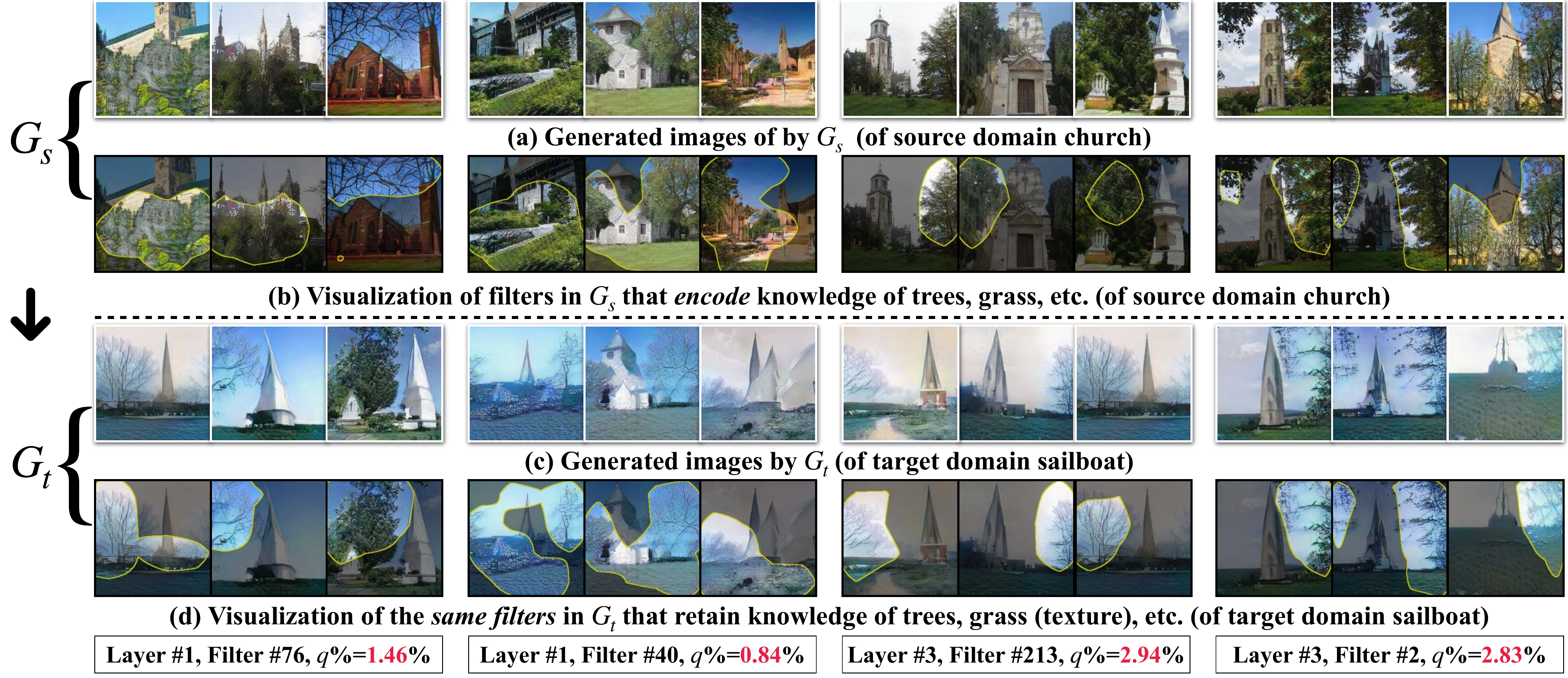}
    \includegraphics[width=0.88\textwidth]{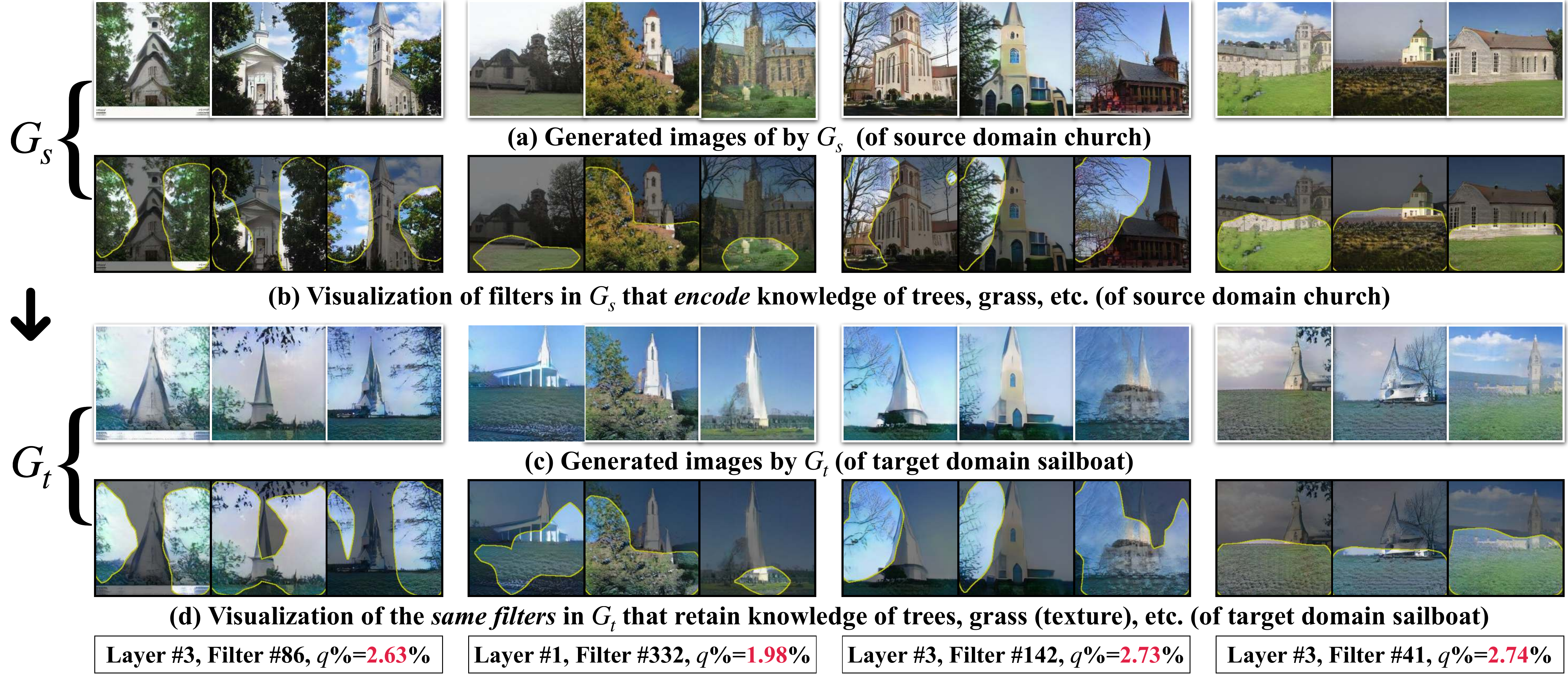}
    \caption{Additional GAN dissection results using EWC.
    Similar to AdAM, in EWC fine-tuning is applied during adaptation from $G_s$ to $G_t$ to update these low importance filters. However, we observe that incompatible knowledge (tree, building, grass) remains in the same filters in $G_t$ after fine-tuning.
    }
    \label{fig:diss_supp_ewc}
\end{figure}

\section{Additional GAN dissection results}
\label{sec-dissect}

\textbf{Additional GAN dissection results for AdAM.}
In Figure {\color{red} 2}, we use IP to estimate the filter importance, therefore we use AdAM as the adaptation method for analysis. 
In this section, we provide additional GAN dissection results as supplement to Figure {\color{red} 2} in the main paper. 

\textbf{Additional GAN dissection results for EWC.}
Since EWC \cite{li2020fig_EWC} also proposed parameter importance estimation method for FSIG adaptation, it is naturally to validate our findings in Sec. {\color{red} 3} with EWC method as well.
In this section, we additionally include EWC for GAN dissection, and we note that we have similar observations as Sec. {\color{red} 3} in the main paper. 

The parameter (filter) importance estimation criteria of AdAM and EWC can be found in Sec.\ref{sec-implementation-detail}.
The detailed additional visualization results can be found in Figure \ref{fig:diss_supp_adam} (AdAM) and Figure \ref{fig:diss_supp_ewc} (EWC).

\section{Ablation Study: 
The impact of high-importance filters}
\label{sec-high-fi-filters}

In the main paper, we emphasize our contributions of investigating the incompatible knowledge transfer, its relationship with the least important filters, and the proposed method to address this unnoticed issue for FSIG.
Besides knowledge truncation, following prior works, we also preserve useful source knowledge for adaptation. 
Specifically, we preserve filters that are deemed important for target adaptation by freezing them. We select the high-importance filters by using a quantile ($t_{h}$, \eg $75\%$) as a threshold. 
See Algorithm \ref{alg:method_algo} for details.
In this section, we conduct a study to show the effectiveness and impact of preserving different amount of filters that are deemed to be most relevant for target adaptation, the results are in Table. Note that we do not prune any filters in this experiment.

\begin{table}[!h]
    \centering
    \caption{
    We preserve different amount of filters during adaptation and evaluate the performance of adapted generators. We do not prune any filters (\ie, $t_{l}=0\%$) in this experiment. The experiment setups are the same as Table {\color{red} 1} in the main paper. $t_{h}$ is the quantile that we start to preserve filters. \eg., if $t_{h}=90\%$, we only preserve $10\%$ filters (see Algorithm {\color{red} 1} for details).
    }
    % \begin{adjustbox}{width=0.85\textwidth}
    \begin{tabular}{ l | c |c | c | c | c | c|c}
    \toprule
         \bm{$t_{h}$}
         & $30\%$
         & $40\%$
         & $50\%$ 
         & $60\%$
         & $70\%$
         & $80\%$
         & $90\%$
        %  & $100\%$ (TGAN \cite{wang2018transferringGAN}) 
         \\ \hline
         Babies
          & $47.55$
          & $47.42$
          & \bm{$46.54$}
          & $49.19$
          & $51.09$
          & $52.00$
          & $67.33$
        %   & $101.58$
          \\
         AFHQ-Cat
          & $90.13$
          & $69.64$
          & $69.13$
          & $65.92$
          & \bm{$57.02$}
          & $57.27$
          & $60.56$
        %   & $64.68$
          \\
    \bottomrule
    \end{tabular}
    % \end{adjustbox}
    \label{table-supp:high_filter}
\end{table}
 As shown in Table \ref{table-supp:high_filter},  varying amounts of filters for preservation do in fact increase performance in different ways.
 In practice, we select $t_{h}=50\%$ for FFHQ $\mapsto$ Babies and $t_{h}=70\%$ for FFHQ $\mapsto$ AFHQ-Cat, and this choice is intuitive: for target domains that are semantically closer to the source, preserving more source knowledge might improve the performance.

\section{Ablation Study: 
Effectiveness of dynamic filter importance estimation}
\label{sec-freeze-adam}
In the main paper, our proposed method includes a dynamic filter importance estimation scheme (denote as ``dynamic''). 
In contrast to prior work \cite{li2020fig_EWC, zhao2022fsig-ip} that evaluate the parameter importance only once before the adaptation stage (denote as ``static''), we regularly estimate the filter importance for target adaptation every certain iterations (\eg 50 iterations in our experiments). 
In this section, we conduct a study to show the effectiveness of the proposed dynamic importance estimation scheme (compared to the ``static'' estimation scheme) and the results are shown in Table \ref{table:ablation_dynamic}.

\begin{table}[h]  
    \caption{
    In this experiment, we study the effectiveness of the proposed dynamic framework for knowledge truncation and preservation. 
    In AdAM, they propose a modulation based method (see details in \cite{zhao2022fsig-ip}) which is static, i.e., they conduct one-time filter importance identification before the main adaptation stage.
    To show the effectiveness of the proposed dynamic framework for FSIG is better than its static counterpart proposed in \cite{zhao2022fsig-ip}, we conduct a study and show results in below. Note that we modify the original AdAM such that it freezes the important parameters rather than modulating them, hence we have a more direct comparison. We use the same experiment setup as Table {\color{red} 1} in the main paper for fair comparison.
    }
   \centering
        \begin{tabular}{l| c c  c c  c c }
        \toprule
        \textbf{Method}
         & \multicolumn{2}{c}{\textbf{FFHQ $\rightarrow$ Babies}}
         & \multicolumn{2}{c}{\textbf{FFHQ $\rightarrow$ Cat}}
         \\ 
         & {FID ($\downarrow$)} & {Intra-LPIPS ($\uparrow$)} & {FID ($\downarrow$)} & {Intra-LPIPS ($\uparrow$)}  \\
                \hline
        TGAN \cite{wang2018transferringGAN} & $101.58$ & $0.517$ & $64.68$ & $0.490$ \\
        EWC \cite{li2020fig_EWC} & $79.93$ & $0.521$ & $74.61$ & $0.587$ \{static\} \\ \hline
        AdAM \cite{zhao2022fsig-ip} \{static, modulation, w/o prune\} & $48.83$ & $0.590$ & $58.07$ & $0.557$ \\
        AdAM \{static, modulation, w/ prune\} \textbf{(Ours)} & \bm{$43.12$} & - & \bm{$53.94$} & -
        \\ \hline
        AdAM \{static, freezing, w/o prune\} & $50.81$ & $0.581$ & $61.60$ & $0.559$ \\
        AdAM \{static, freezing, w/ prune\} \textbf{(Ours)} & $46.87$ & - &  $57.56$ & - \\ \hline
        \textbf{Ours} \{dynamic, freezing,  w/prune\} & \bm{$39.39$} & \bm{$0.608$} & \bm{$53.27$} & $0.569$ \\
        \bottomrule
        \end{tabular}
    \label{table:ablation_dynamic}
\end{table}

\section{Ablation Study: 
Additional importance measurement}
\label{sec-supp:class_saliency}

Evaluating the importance of weights in generative tasks is still underexplored. 
In the main paper, we follow some prior works \cite{li2020fig_EWC, achille2019task2vec} to use the Fisher Information (FI) as the measurement for importance estimation and obtain competitive performance across different datasets (See Table {\color{red} 1} in the main paper).
Nevertheless, there could be different ways to evaluate how well the obtained weights given the adaptation task.
In literature, Class Salience \cite{simonyan2013deep} (CS) is used as a tool to estimate which area/pixels of a given input image stand out for a specific classification decision, and it is similar to FI that leverages the gradient information.
Therefore, we note that CS could have a connection with FI as they both use the knowledge encoded in the gradients for knowledge importance estimation.

We perform an experiment to replace FI with CS in filter importance measurement and compare with prior works. Note that, in \cite{simonyan2013deep}, CS is computed w.r.t. input image pixels. 
To make CS suitable for our problem, we modify it and compute CS 
% (\ie, absolute value of the gradient in our problem)
w.r.t. different filters by averaging the importance of all parameters within a filter to calculate the importance of that filter. 
After that, the same as main paper, we preserve the filters that are deemed to be salient for target adaptation by freezing them, prune the filters with least CS value and fine-tune the rest of filters, and we regularly evaluate the filter importance.
The results in Table \ref{table:cs-fi} are obtained with our proposed method using FI and CS as the measurements for filter importance estimation.

\begin{table}[h]  
    \caption{
    {
    % \color{blue}
    In this experiment, we apply different measurements (\ie, FI and CS) for filter importance estimation and compare with state-of-the-art methods. 
    The experiment setup is the same as Table {\color{red} 1} in the main paper.
    We evaluate the performance under different source $\rightarrow$ target adaptation setups where source and target domains have different proximity.}
    }
   \centering
        \begin{tabular}{l| c c  cc }
        \toprule
        \textbf{Method}
         & \multicolumn{2}{c}{\textbf{FFHQ $\rightarrow$ Babies}}
         & \multicolumn{2}{c}{\textbf{FFHQ $\rightarrow$ Cat}}
         \\ 
         & {FID ($\downarrow$)} & {Intra-LPIPS ($\uparrow$)} & {FID ($\downarrow$)} & {Intra-LPIPS ($\uparrow$)}  \\
                \hline
        % Class Salience \cite{simonyan2013deep}  & 52.46 & 0.582 & 61.68 & 0.556 \\
        TGAN \cite{wang2018transferringGAN} & 101.58 & 0.517 & 64.68 & 0.490 \\
        EWC \cite{li2020fig_EWC} & $79.93$ & $0.521$ & $74.61$ & $0.587$ \\
        AdAM \cite{zhao2022fsig-ip} & $48.83$ & $0.590$ & $58.07$ & $0.557$ \\ \hline
        \textbf{Ours} w/ Class Salience  & $39.95$ & $0.607$ & $56.46$ & $0.574$ \\
        \textbf{Ours} w/ Fisher Information & \bm{$39.39$} & \bm{$0.608$} & \bm{$53.27$} & $0.569$ \\
        \bottomrule
        \end{tabular}
    \label{table:cs-fi}
\end{table}

% \color{blue}
As can be observed in Table \ref{table:cs-fi}, our proposed method can be applied with different measurements for importance estimation.
Meanwhile, we note that a good FID score does not necessarily infer to a good diversity of generated images from the obtained generator \cite{webster2019detecting_overfit_fid}.
Compared to prior works, our proposed method can consistently achieve a \textbf{good trade-off} of performance (between FID and intra-LPIPS) of the adapted models across different adaptation setups.

\section{Additional qualitative results}
\label{sec-quality}
{\bf Few-shot adaptation with additional setups.}
We provide qualitative results with additional adaptation setups to show the generalizability of our proposed method.
Specifically, we visualize the generated images before and after adaptation. We use the same experiment setups as in the main paper and 10-shot target samples for adaptation.
As results below, since our proposed method reliably removes the filters that are deemed to be incompatible with the target domains,  \textbf{there is not much incompatible knowledge transferred to the generated images}. 
Meanwhile, we show that the generated images are diverse and high fidelity, as we also preserve knowledge important for target adaptation:
\begin{itemize}
    \item \textbf{FFHQ $\mapsto$ Sketches:} Figure \ref{fig:supp-sketches}
    
    \item \textbf{FFHQ $\mapsto$ Sunglasses:} Figure \ref{fig:supp-sunglasses}
    
    \item \textbf{FFHQ $\mapsto$ Metfaces:} 
    Figure \ref{fig:supp-metfaces}
    
    % \item \textbf{FFHQ $\mapsto$ Otto Dix 's Paintings:}
    
    \item \textbf{FFHQ $\mapsto$ Raphael's Paintings:} Figure \ref{fig:supp-raphael}
    
    \item \textbf{FFHQ $\mapsto$ Amedeo-Modigliani’s Paintings:}
    Figure \ref{fig:supp-amedeo}
    
    % \textbf{\item FFHQ $\mapsto$ AFHQ-Dog:}
    % Figure
    
    % \textbf{\item FFHQ $\mapsto$ AFHQ-Wild:}
    % Figure
    
    \textbf{\item LSUN Church $\mapsto$ Haunted House:} 
    Figure \ref{fig:supp-haunted}
    
    % \textbf{\item LSUN Church $\mapsto$ Van Goah Houses:}
    % Figure
    
    \textbf{\item LSUN Church $\mapsto$ Palace:}
    Figure \ref{fig:supp-palace}

    % \textbf{\item LSUN Cars $\mapsto$ Wrecked Cars:}
    % Figure
\end{itemize}

\begin{figure}[h]
    \centering
    \includegraphics[width=0.91\textwidth]{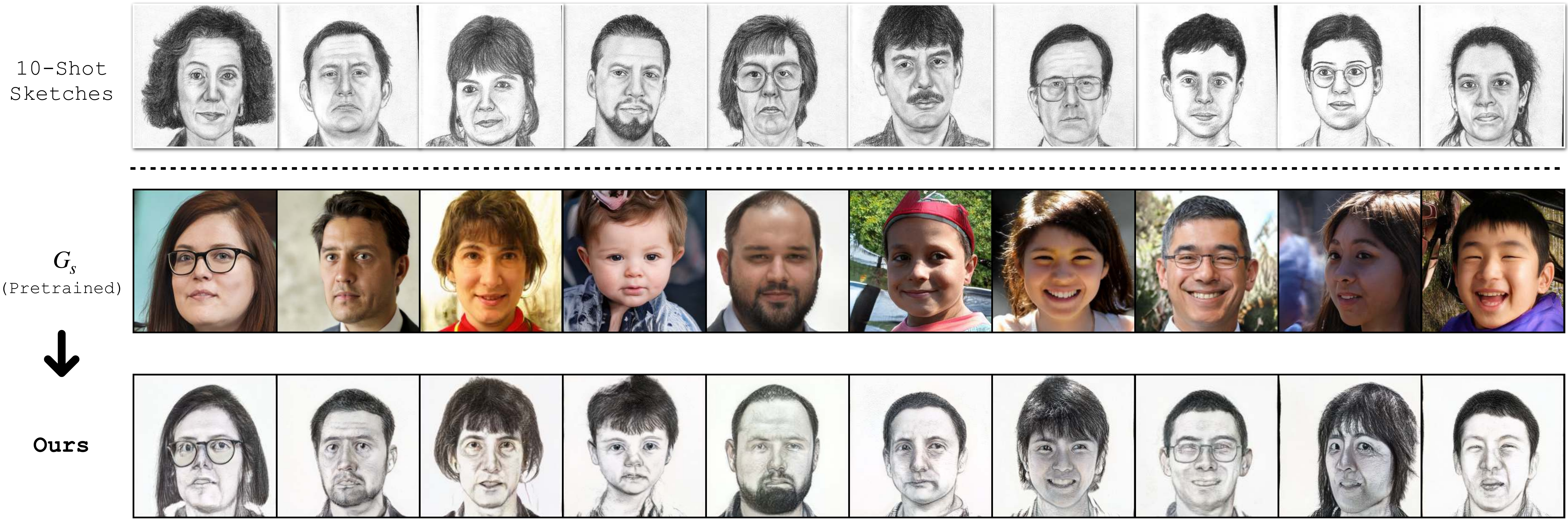}
    \caption{Additional FSIG results with FFHQ $\mapsto$ Sketches.}
    \label{fig:supp-sketches}
\end{figure}

\begin{figure}[h]
    \centering
    \includegraphics[width=0.91\textwidth]{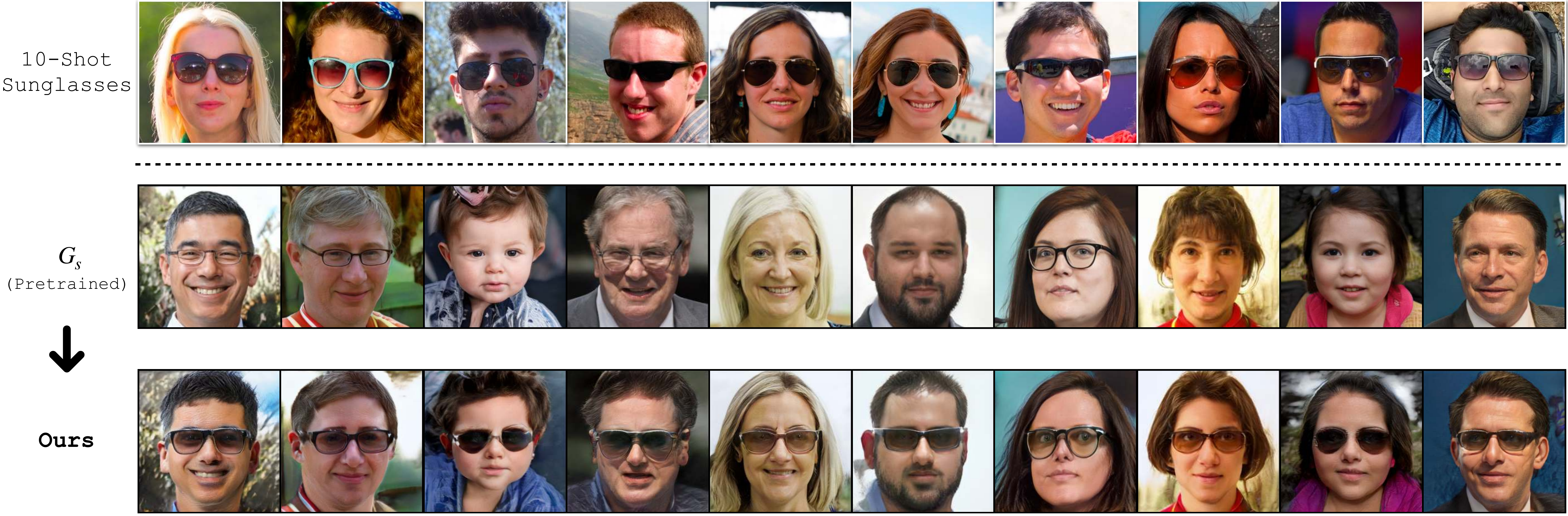}
    \caption{Additional FSIG results with FFHQ $\mapsto$ Sunglasses.}
    \label{fig:supp-sunglasses}
\end{figure}

\begin{figure}[h]
    \centering
    \includegraphics[width=0.91\textwidth]{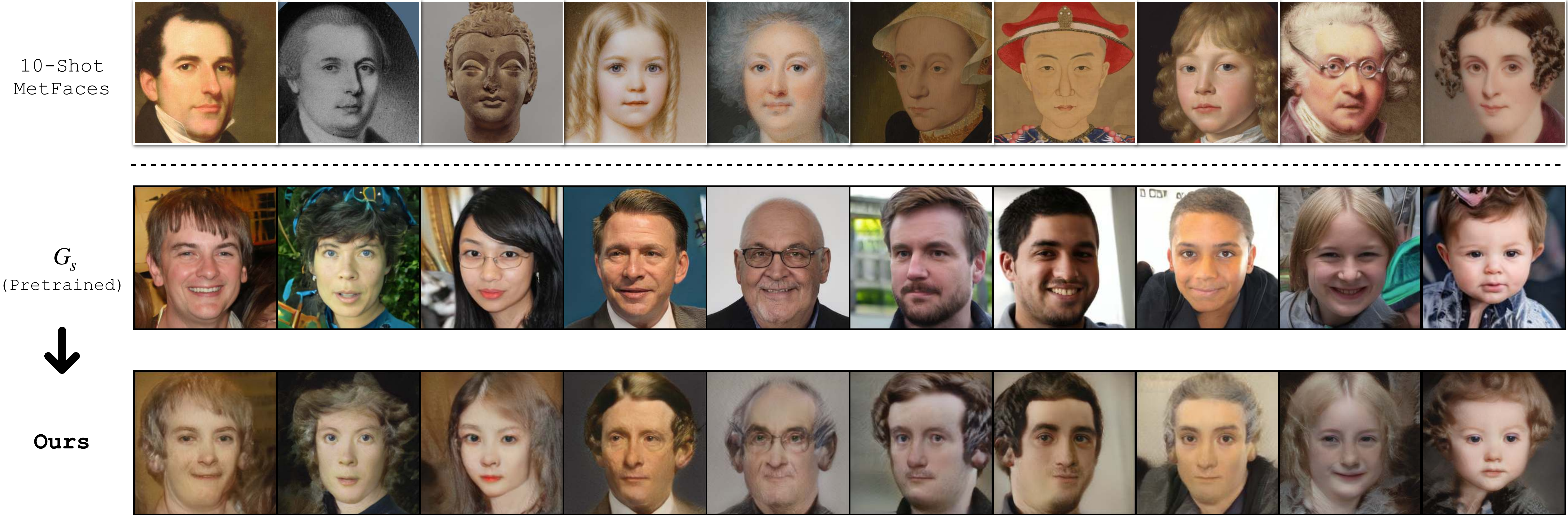}
    \caption{Additional FSIG results with FFHQ $\mapsto$ MetFaces.}
    \label{fig:supp-metfaces}
\end{figure}

\begin{figure}[h]
    \centering
    \includegraphics[width=0.91\textwidth]{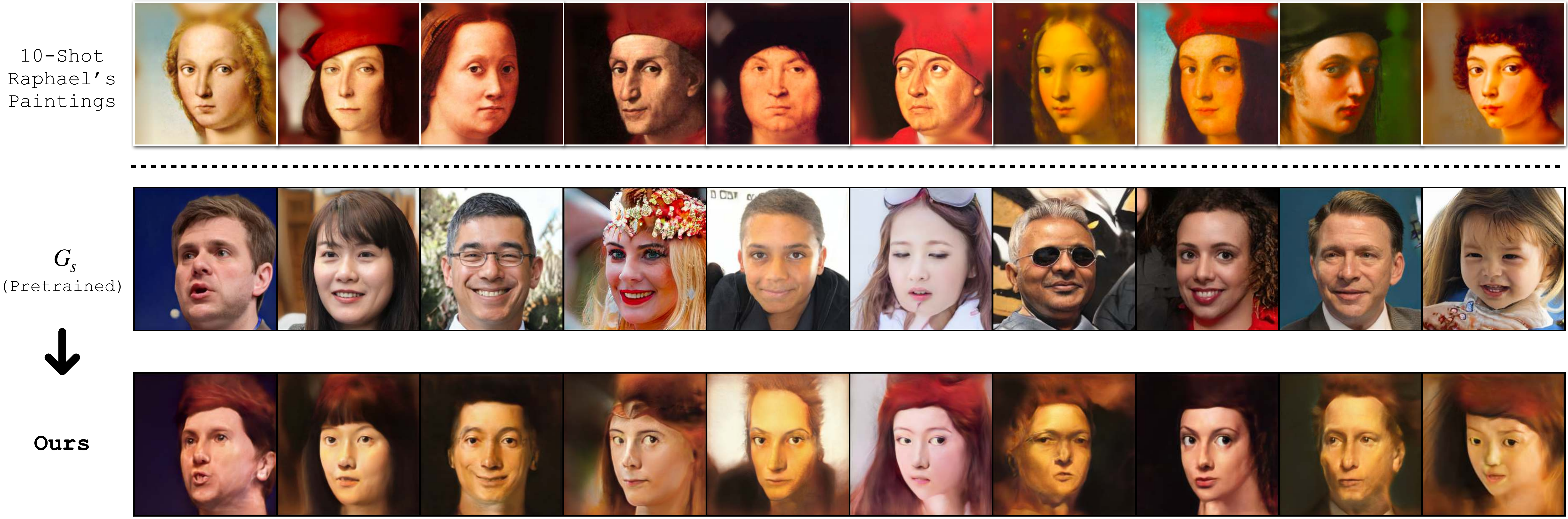}
    \caption{Additional FSIG results with FFHQ $\mapsto$ Raphael's paintings.}
    \label{fig:supp-raphael}
\end{figure}

\begin{figure}[h]
    \centering
    \includegraphics[width=0.91\textwidth]{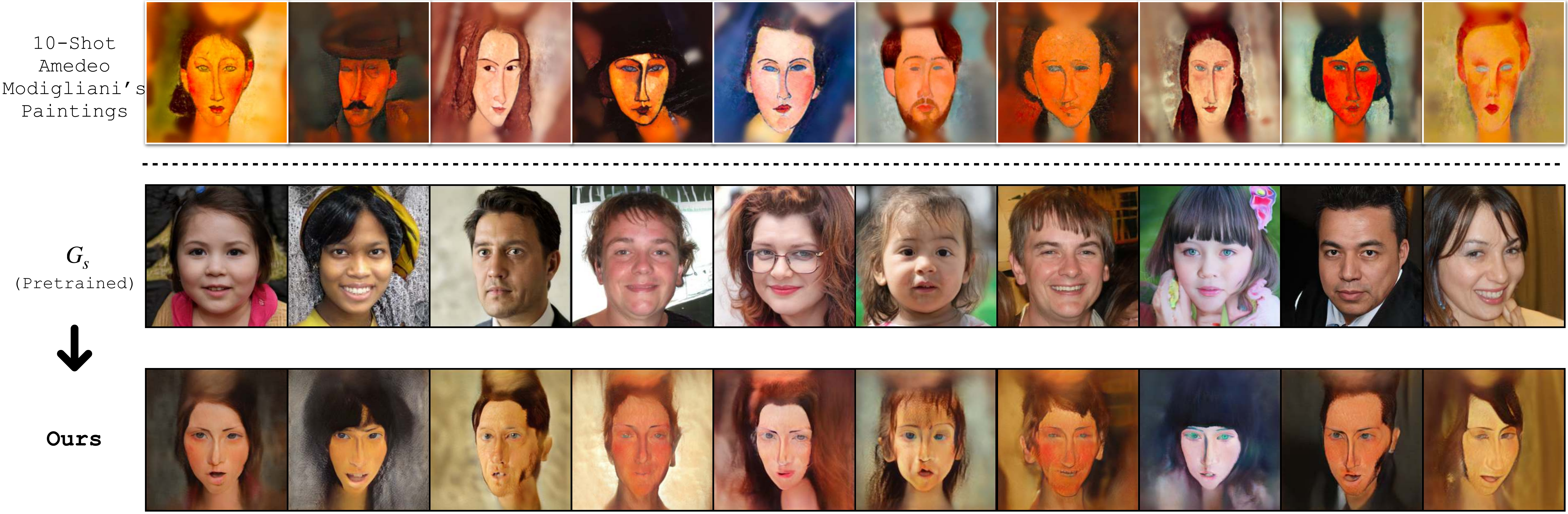}
    \caption{Additional FSIG results with FFHQ $\mapsto$ Amedeo-Modigliani 's paintings.}
    \label{fig:supp-amedeo}
    \vspace{-3mm}
\end{figure}

\begin{figure}[h]
    \centering
    \includegraphics[width=0.91\textwidth]{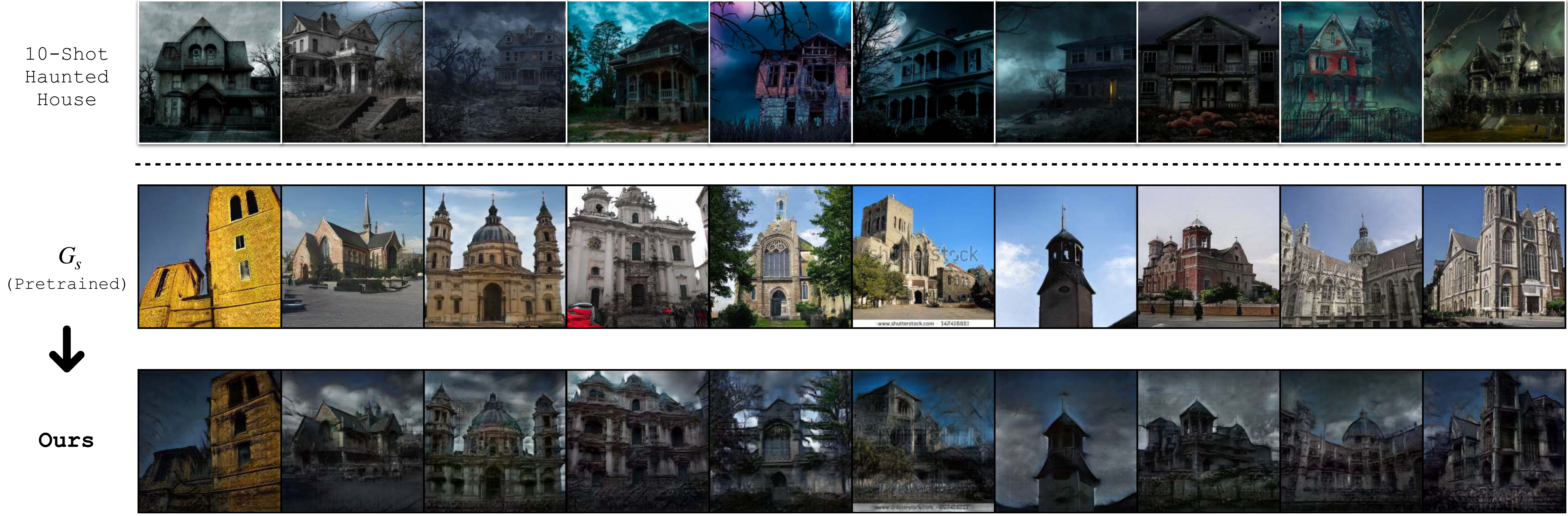}
    \caption{Additional FSIG results with Church $\mapsto$ Haunted Houses.}
    \label{fig:supp-haunted}
\end{figure}

\begin{figure}[h]
    \centering
    \includegraphics[width=0.91\textwidth]{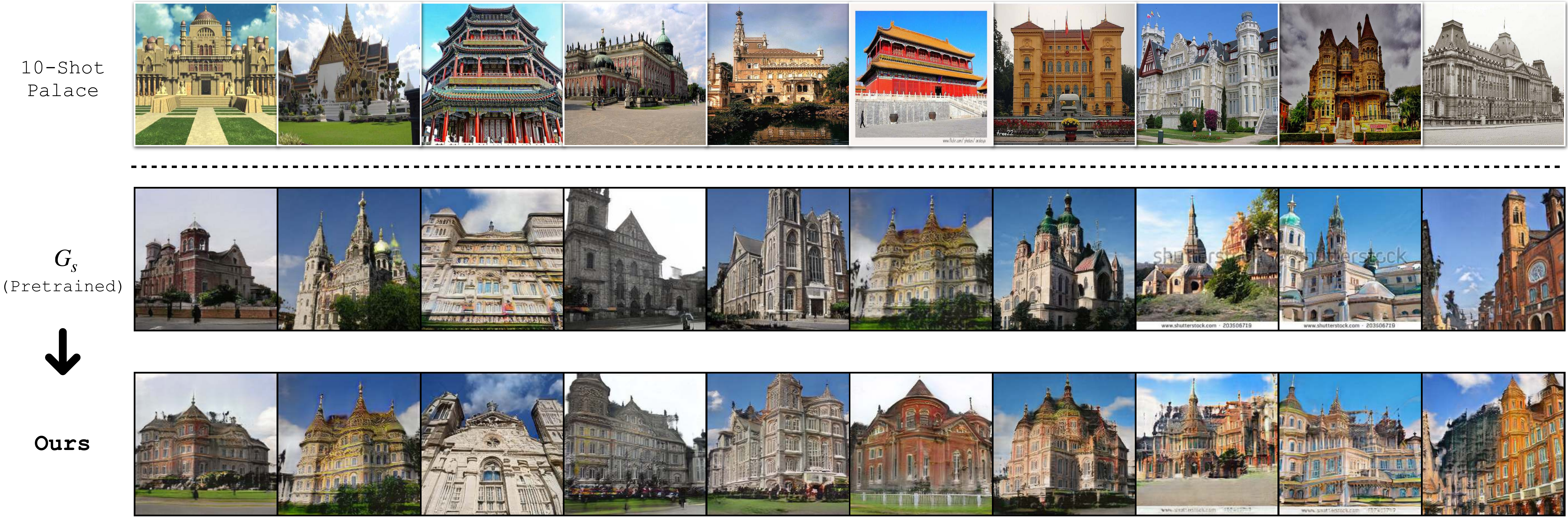}
    \caption{Additional FSIG results with Church $\mapsto$ Palace}
    \label{fig:supp-palace}
\end{figure}

{\bf Interpolation in the latent space.} 
In our methods, the MLP layers in the generator are not tuned. In Fig. \ref{r1} , we 
show the semantic manipulation results of our target model that it facilitates editing applications. We follow GenDA \cite{yang2021one-shot-adaptation} to linearly interpolate between two different latent codes after few-shot adaptation and visualize the intermediate generated images, which are still of high quality.

\begin{figure}[h]
    \centering
    \includegraphics[width=0.91\textwidth]{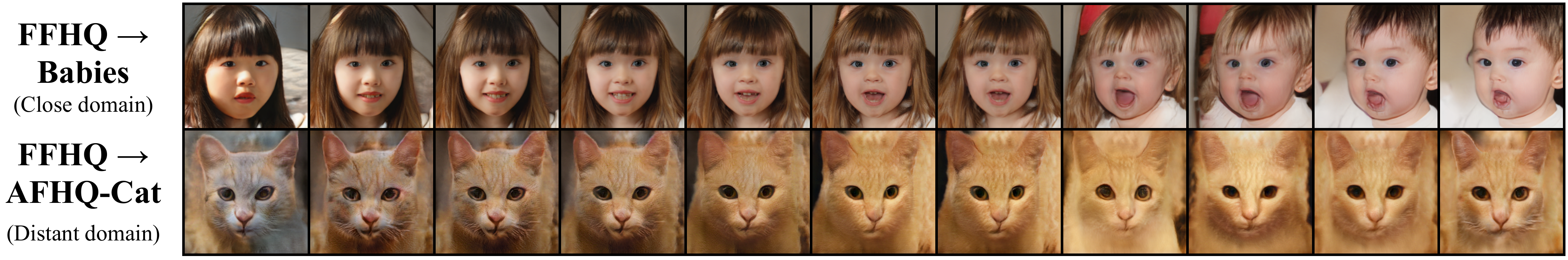}
    \caption{
     Visualization of the generated images using linear interpolation between two latent codes after adaptation.}
    \label{r1}
\end{figure}

{\bf Comparison with SOTA on additional target domains.} The discussion and results are in Figure \ref{r2_q3}.

\begin{figure}[h]
    \centering
    \includegraphics[width=\textwidth]{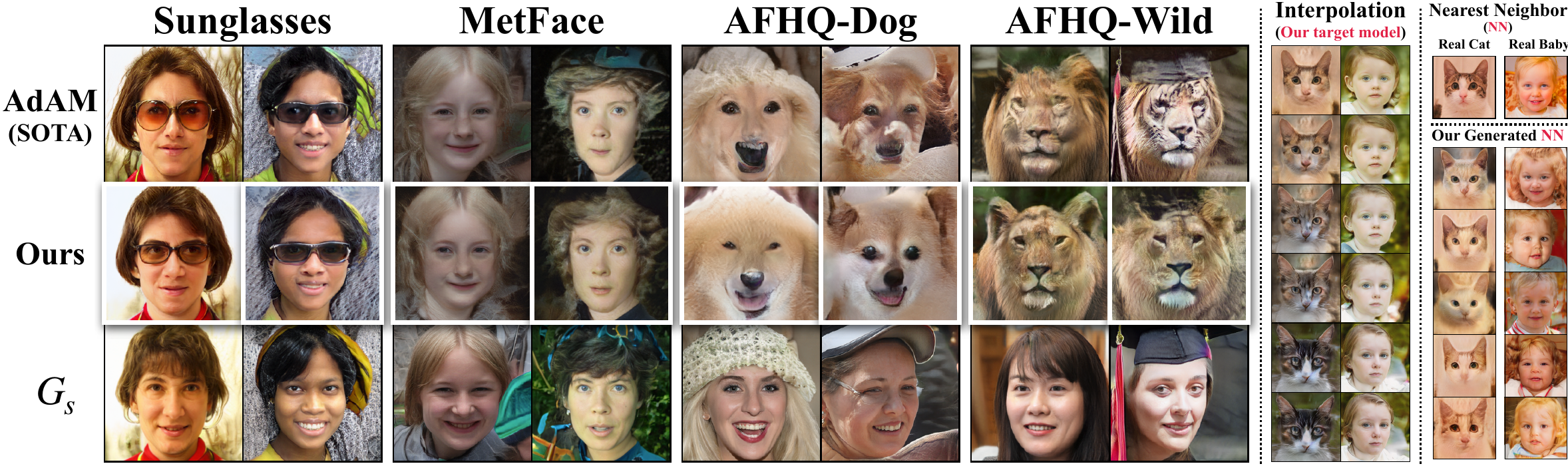}
    % \captionsetup{font={stretch=0.9}}
    \caption{
    Additional visualization and comparison results. 
    {\bf Left:} Comparison with SOTA on additional target domains (Sunglasses, MetFaces, etc.), FFHQ is the source domain.
    {\bf Mid:} Similar to Figure \ref{r1}, we provide additional visualized images by using linearly interpolated latent codes.
    {\bf Right:}
    For a selected target image, we visualized the nearest neighbour of that target image. To this end, we generate 5,000 images after adaptation, and assign images with smallest LPIPS distance \cite{zhang2018lpips} to each of the target image.
    \textbf{Best viewed in color and zooming in.}
    }
    \label{r2_q3}
\end{figure}

\section{Additional quantitative results}
\label{sec-quantity}

\textbf{FSIG with different number of shots.}
In the main paper, we mainly focus on the adaptation setup where we have only 10-shot samples for adaptation. 
In this section, we comprehensively evaluate the proposed method given different number of target training samples. 
The results are in Table \ref{table:supp-fid-shot}.

\begin{table*}[h]  
    \renewcommand{\arraystretch}{0.95}
    \centering
    \caption{
    FID ($\downarrow$) with respect to the number of shots for adaptation. We use the same experiment setups as Table {\color{red} 1} in the main paper. 
    }
    \begin{adjustbox}{width=0.92\textwidth,center}
    \begin{tabular}{p{3cm}<{\centering} | p{2.2cm}<{\centering}  | cccccccccc }
    \toprule
    \textbf{Number of Samples}
    & \textbf{Domain}
    & TGAN \cite{wang2018transferringGAN}
    & TGAN+ADA \cite{karras2020ADA}
    & EWC \cite{li2020fig_EWC}
    & AdAM \cite{zhao2022fsig-ip}
    & \textbf{Ours} \\
    \midrule
    1-shot  & & $172.49$ & $188.84$ & $104.5$ & $77.71$ & \bm{$74.34$}\\
    5-shot  & & $108.65$ & $105.19$ & $88.51$ & $52.85$ & \bm{$43.53$}  \\
    10-shot & & $101.58$ & $102.98$ & $79.93$ & $48.83$ & \bm{$39.39$} \\
    25-shot & 
    & $54.83$ & $56.66$ & $44.67$ & $27.77$ & \bm{$25.13$} \\
    50-shot & FFHQ $\mapsto$ & 
    $48.39$ & $52.94$ & $39.32$ & $24.69$ & \bm{$22.46$} & 
      \\
    100-shot & Babies & $39.04$ & $45.71$ & $34.49$ & $19.63$ & \bm{$18.85$} \\
    200-shot & & $33.65$ & $38.84$ & $32.65$ & $17.06$ & \bm{$15.71$} \\
    500-shot & & $27.21$ & $26.31$ & $28.11$ & $16.17$ & \bm{$13.67$} \\
    All & & $25.63$ & $25.47$ & $24.57$ & $13.59$ & \bm{$12.34$} \\
    \bottomrule
    \end{tabular}
    \end{adjustbox}
    \begin{adjustbox}{width=0.92\textwidth,center}
    \begin{tabular}{p{3cm}<{\centering} | p{2.2cm}<{\centering} | cccccccc}
    \toprule
    \textbf{Number of Samples}
    & \textbf{Domain}
    & TGAN \cite{wang2018transferringGAN}
    & TGAN+ADA \cite{karras2020ADA}
    & EWC \cite{li2020fig_EWC}
    & AdAM \cite{zhao2022fsig-ip}
    & \textbf{Ours} \\
    \midrule
    1-shot   & & $125.52$ & $125.81$ & $139.11$ & $118.25$ & \bm{$117.69$} \\
    5-shot   & & $90.24$ & $86.94$ & $136.65$ & $79.53$ & \bm{$74.47$}\\
    10-shot  & & $64.68$ & $80.16$ & $74.61$ & $58.07$  &  \bm{$53.27$} \\
    25-shot  & & $40.52$ & $48.61$ & $56.23$ & $32.38$  & \bm{$31.85$}  \\
    50-shot  & FFHQ $\mapsto$ & $33.87$ & $35.76$  & $43.58$ & $26.43$ & \bm{$24.67$} \\
    100-shot & AFHQ-Cat & $27.78$ & $28.16$ & $36.93$  & $21.50$ & \bm{$19.60$} \\
    200-shot &  & $24.73$ & $26.78$ & $33.43$ & $19.79$  & \bm{$18.02$} & \\
    500-shot & & $20.25$ & $19.01$ & $32.73$ & $16.80$  & \bm{$13.56$} \\
    All  & & $10.52$ & $9.56$ & $18.76$ & $6.52$ & \bm{$6.22$} \\
    \bottomrule
    \end{tabular}
    \end{adjustbox}
\label{table:supp-fid-shot}
\end{table*}

\textbf{Evaluation on additional dataset.}
We note that we have evaluate our method and compare with SOTA methods comprehensively in the main paper, see Table {\color{red} 1} and Figure {\color{red} 4}. 
Here, we include additional results of FFHQ $\mapsto$ Sketches \cite{wang2008cuhk_sketches}. The results are included in Table \ref{table-supp:sketches}.

\begin{table}[!h]
    \centering
    \caption{
    We additionally evaluate the proposed method in FFHQ $\mapsto$ Sketches setup. 
    We use the same experiment setup as Table {\color{red}1} in the main paper.
    Compared to prior state-of-the-art methods \cite{li2020fig_EWC, kingma2014adam}, our proposed method can achieve better FID by a large margin with comparable diversity. 
    Meanwhile, we also note that the entire Sketch domain \cite{wang2008cuhk_sketches} contains only $\sim$300 images, which could be not very stable for evaluation and we only include the results in Supplementary for reference.
    }
    % \begin{adjustbox}{width=0.85\textwidth}
    \begin{tabular}{ l | c |ccccccccc}
    \toprule
         \textbf{Metric} & \textbf{Domain} 
         & TGAN \cite{wang2018transferringGAN} 
         & TGAN+ADA \cite{karras2020ADA} 
         & EWC \cite{li2020fig_EWC} 
         & CDC \cite{ojha2021fig_cdc}
         & AdAM \cite{zhao2022fsig-ip} 
         & \textbf{Ours} \\ \hline
          FID ($\downarrow$) & FFHQ $\mapsto$ 
          & $53.42$
          & $66.99$ 
          & $71.25$
          & $45.67$
          & $45.03$ 
          & \bm{$39.29$} \\
          Intra-LPIPS ($\uparrow$) & Sketches 
          & $0.394$
          & $0.414$
          & $0.421$
          & $0.453$
          & $0.459$ 
          & $0.442$ \\
    \bottomrule
    \end{tabular}
    % \end{adjustbox}
    \label{table-supp:sketches}
\end{table}

\textbf{Additional evaluation metric.}
We note that we evaluate our proposed method via different quantiative metrics (FID, Intra-LPIPS) across six different datasets, and achieve competitive performance compared to prior works.
Here we additionally provide the results of Kernel Inception Score (KID \cite{binkowski2018kid_score}), which is supplement of FID in our work, to evaluate the difference between abundant generated images and the entire target domain. The results are in Table \ref{table-supp:kid}. We show that our proposed method can achieve comparable KID score with other SOTA FSIG methods.

{\bf Additional Ablation Studies.}
{\bf Table \ref{r2_q1} 1) \& 3): We conduct experiments and show that our proposed dynamic selection is indeed \textit{important and compatible} with pruning.}
{\bf Table \ref{r2_q1} 5) \& 6)}: We randomly re-initialize the filters with the lowest importance, and we show that those filters may not learn knowledge properly due to less iterations in few-shot adaptation.
Therefore, we propose to prune lowest important filters.
{\bf Table \ref{r2_q1} 7)}: We show the results of dynamic scheme with modulation method \cite{zhao2022fsig-ip}, the performance is comparable with the freezing scheme. In our paper, we adopt freezing which is simple to implement.
\begin{table}[h]  
    \centering
    % \captionsetup{font={stretch=0.9}}
    \caption{
    % \footnotesize
    FID ($\downarrow$) with the same hyper-parameters as {\color{RubineRed} Tab. S2}.
    We follow {\color{RubineRed} Tab.S2} to evaluate the filter importance for target adaptation: only once for \textit{static} and periodically for \textit{dynamic}. 
    }
   \begin{adjustbox}{width=0.83\columnwidth,center}
        \begin{tabular}{l| c c  c c  c c }
        \toprule
        \textbf{Method} {\color{RubineRed}(the same experiment setup as Tab. S2)}
        & {\textbf{FFHQ $\rightarrow$ Babies}}
        & {\textbf{FFHQ $\rightarrow$ Cat}}
         \\ \hline
        \textbf{1) Ours} \{static, no freezing,  w/ prune\} & $56.62$ & $64.25$ \\ 
        \textbf{2) Ours} \{static, w/ freezing, w/ prune\} {\color{RubineRed}(Tab.S2)} & $46.87$ &  $57.56$ \\
        \textbf{3) Ours} \{dynamic, no freezing,  w/ prune\} & 
        $51.08$ & $59.71$ \\
        \textbf{4) Ours} \{dynamic, w/ freezing,  w/ prune\} {\color{RubineRed}(Tab. S2 \& Tab. 1)}  & \bm{$39.39$} & \bm{$53.27$} \\\hline
        \textbf{5) Ours} \{static,  w/ freezing,  w/ random re-initialize\} & $50.54$ & $59.93$ \\
        \textbf{6) Ours} \{dynamic, w/ freezing,  w/ random re-initialize\} & $41.67$ & $55.21$ \\\hline
        \textbf{7) Ours} \{dynamic, w/ modulation,  w/ prune\} & $39.97$ & $53.81$ \\
        \bottomrule
        \end{tabular}
    \end{adjustbox}
    \label{r2_q1}
\end{table}

\begin{table}[!h]
    \centering
    \caption{
    Besides the FID and Intra-LPIPS results in the main paper, we additionally include KID ($\downarrow$) score of different methods.
    We use the same experiment setup as Table {\color{red}1} in the main paper. 
    Following prior works \cite{karras2020ADA, chai2021ensembling}, we scale the KID values with $10^3$.
    }
    % \begin{adjustbox}{width=0.99\textwidth}
    \begin{tabular}{l|cccccccccc}
    \toprule
         \textbf{Domain} & TGAN \cite{wang2018transferringGAN} & TGAN+ADA \cite{karras2020ADA}
         & EWC \cite{li2020fig_EWC} & CDC \cite{zhao2022dcl} & AdAM \cite{zhao2022fsig-ip} & \textbf{Ours} \\ \hline
         FFHQ $\mapsto$ Babies  & $60.91$ & $64.84$ & $54.94$ & $47.45$ & $32.74$ & \bm{$29.43$}\\
         FFHQ $\mapsto$ AFHQ-Cat & 
         $46.03$ & $64.54$ & $57.79$ & $192.38$ & $35.53$ & \bm{$35.16$} 
         \\
    \bottomrule
    \end{tabular}
    % \end{adjustbox}
    \label{table-supp:kid}
\end{table}

\section{Training longer leads to severe overfitting of existing FSIG methods}
\label{sec-train-longer}
% As discussed in Sec. {\color{red} 6} in the main paper, to``remove'' the knowledge that is incompatible to the target domain, one simple way could be fine-tuning \textit{for more iterations} by Eqn. {\color{red} 1} in the main paper (\ie, GAN loss). 
%
In this section, we conduct a study
to show that issue of incompatible knowledge \textit{cannot be addressed by fine-tuning for more iterations} by Eqn. {\color{red} 1} in the main paper (\ie, GAN loss), as with more iterations in fine-tuning would lead to severe overfitting as also  pointed out in previous work \cite{zhao2022dcl}.
%and show this does not work well and it results in severe mode collapse for existing FSIG methods.
Specifically, we follow the experiment setup in the main paper (\eg Figure {\color{red} 1} and Figure {\color{red} 2} as we analyze the incompatible knowledge transfer in that section) and evaluate the diversity of generated images at different adaptation steps. The results are in Table \ref{table-supp:train-longer} and it clearly shows the diversity of generated images gradually collapse to the few-shot target samples. 
We conjecture that this is due to $\mathcal{L}_{adv}$ where the generatore $G_{t}$ is encouraged to replicate few-shot target samples to fool the discriminator. 
Therefore, based on the above observations, it is important to remove the knowledge incompatible to the target domain before mode collapse becomes severe.

\begin{table}[!h]
    \centering
    \caption{
    \textbf{Intra-LPIPS of generated images during adaptation} (see details of Intra-LPIPS in Algorithm \ref{intra-lpips-algo}).  
    The issue of incompatible knowledge cannot be addressed by training longer iterations. 
    %Naturally, if we train longer iterations for adaptation, we might be able to ``remove'' the incompatible knowledge after adaptation.
    In particular, we empirically observe that training longer iterations will lead to severe overfitting for different existing FSIG methods, including early baseline method \cite{wang2018transferringGAN} and recent state-of-the-art methods \cite{li2020fig_EWC, zhao2022fsig-ip}.
    Here we evaluate the diversity of generated images of different methods using Intra-LPIPS over 10-shot target samples. Lower value indicates that the generated images are less diverse and collapsing to few-shot target sample.
    The experiment setups are the same as Table {\color{red} 1} in the main paper. 
    }
    % \begin{adjustbox}{width=0.7\textwidth}
    \begin{tabular}{ l | p{2cm}<{\centering} | ccccccccc}
    \toprule
         \textbf{Method} & \textbf{Domain}
         & iter-0 & iter-500 & iter-750 & iter-1000 & iter-1250 \\ \hline
         TGAN \cite{wang2018transferringGAN} & 
         & $0.678$ & $0.563$ & $0.507$ & $0.436$
         & $0.394$ \\
         EWC \cite{li2020fig_EWC} & Church $\mapsto$ & $0.679$ & $0.641$ & $0.611$ & $0.599$ & $0.583$ \\
         AdAM \cite{zhao2022fsig-ip} & Sailboat 
         & $0.693$ & $0.611$ & $0.546$ & $0.512$ & $0.484$
         \\
         Ours & & $0.683$ & $0.605$ & $0.559$ & $0.528$ & $0.509$ \\
    \bottomrule
    \end{tabular}
    % \end{adjustbox}
    % \begin{adjustbox}{width=0.7\textwidth}
    \begin{tabular}{ l | p{2cm}<{\centering} | ccccccccc}
    \toprule
         \textbf{Method} & \textbf{Domain}
         & iter-0 & iter-500 & iter-750 & iter-1000 & iter-1250 \\ \hline
         TGAN \cite{wang2018transferringGAN} 
         & & $0.664$ & $0.620$ & $0.563$ & $0.545$ & $0.517$
         \\
         EWC \cite{li2020fig_EWC} 
         & FFHQ $\mapsto$ 
         & $0.665$ & $0.656$ & $0.638$ & $0.622$ & $0.618$ \\
         AdAM \cite{zhao2022fsig-ip} & AFHQ-Cat & $0.670$ & $0.658$ & $0.615$ & $0.585$ & $0.557$
         \\
         Ours  &  & $0.667$ & $0.655$ & $0.635$ & $0.605$ & $0.592$
         \\
    \bottomrule
    \end{tabular}
    % \end{adjustbox}
    \label{table-supp:train-longer}
\end{table}

% \section{Generalization to different GAN architectures}

% -------------------------- %
\section{Future works}
\label{future-works}
We review current SOTA approaches for FSIG in this work and investigate the transfer of incompatible knowledge after adaptation, which significantly reduces the realism of generated images from adapted generators.
We also provide a thorough evaluation of alternative approaches using various adaption scenarios.
We believe our proposed method can also be applied to other generative models and adaptation setups, as it works by adaptively eliminating filters that are incompatible with the target domain while maintaining knowledge crucial to the target domain, which is robust to different generative model architectures and training strategies.
In future research, we are also interested in exploring the knowledge transfer for other generative models, including Diffusion Models \cite{rombach2021highresolution}.
Furthermore, the effects of transferring incompatible knowledge on downstream tasks is also a interesting problem to study.

\section{Broader impact}
\label{broader-impact}

\subsection{Potential social and ethical impact}
Throughout the paper, we demonstrate effective target adaption results using extremely small target training sample(s).
Although we have achieved new state-of-the-art performance across various FSIG setups, we caution that because the FSIG adaptation of our method is lightweight, it could be quickly and cheaply applied to a real person in practice, there may be potential social and ethical issues if it is used by malicious users.
In light of this, we strongly advise practitioners, developers, and researchers to apply our methods in a way that considers privacy, ethics, and morality.

% \textbf{Adaptation to a particular person.}
% {\color{red} least important experiment}

\subsection{Amount of computation and CO$_{2}$ emission}
% {\color{red} copy from AdAM, to be edited}
\label{sec-supp:amount_compute}
Our work includes a large number of experiments, and we have provided thorough data and analysis when compared to earlier efforts.
In this section, we include the amount of compute for different experiments along with CO$_{2}$ emission. 
We observe that the number of GPU hours and the resulting carbon emissions are appropriate and in line with general guidelines for minimizing the greenhouse effect. 
Compared to existing works in computer vision tasks that adopt large-scale pretraining  \cite{he2020moco, rombach2021highresolution} and consume a massive amount of energy, our research is not heavy in computation. 
We summarize the estimated results in Table \ref{table-supp:compute}.

\begin{table}[!ht]
\caption{
Estimation of amount of compute and CO$_{2}$ emission in this work. 
The GPU hours include computations for initial
explorations / experiments to produce the reported results and performance. 
CO$_{2}$ emission values are
computed using Machine Learning Emissions Calculator: \url{https://mlco2.github.io/impact/} 
\cite{lacoste2019quantifying_co2}.
}
  \centering
  \begin{adjustbox}{width=0.98\textwidth}
  \begin{tabular}{l|c|c|c}
  \toprule
\textbf{Experiments} &\textbf{Hardware Platform } &\textbf{GPU Hours (h)} &\textbf{Carbon Emission (kg)} \\ \toprule
Main paper : Table {\color{red} 1} (Repeated three times) & NVIDIA A100-PCIE (40 GB) &312 & 33.7 \\ 
Main paper : Figure {\color{red} 1} & NVIDIA A100-PCIE (40 GB) & 64 & 6.91 \\ 
Main paper : Figure {\color{red} 2} & NVIDIA A100-PCIE (40 GB) & 112 & 12.1 \\
Main paper : Figure {\color{red} 4} \& Figure {\color{red} 5} & NVIDIA A100-PCIE (40 GB) & 87 & 9.4 \\ \hline
Supplementary : Additional Experiments \& Analysis & NVIDIA A100-PCIE (40 GB) & 58 & 6.26 \\ 
Supplementary : Ablation Study & NVIDIA A100-PCIE (40 GB) &16 & 1.73 \\ \hline
Additional Compute for Hyper-parameter tuning & NVIDIA A100-PCIE (40 GB) & 24 & 2.59 \\ \hline
\textbf{Total} &\textbf{--} &\textbf{673} &\textbf{72.69} \\
\bottomrule
\end{tabular}
\end{adjustbox}
\label{table-supp:compute}
\end{table}

% -------------------------- %

% %%%%%%%%% REFERENCES
% {\small
% \bibliographystyle{ieee_fullname}
% \bibliography{egbib}
% }

% \end{document}

}

\end{document}